%% file: main.tex
\documentclass{article}
\usepackage{fancyhdr}

\usepackage{report}

\usepackage{subfigure}
\usepackage{soul}
\usepackage[utf8]{inputenc} 
\usepackage[T1]{fontenc}    
\usepackage{hyperref}       
\usepackage{url}            
\usepackage{booktabs}       
\usepackage{amsfonts}       
\usepackage{fancyhdr}       

\usepackage{nicefrac}       
\usepackage{microtype}      
\usepackage{graphicx}
\usepackage{pifont}
\usepackage{multirow}
\usepackage{CJKutf8}
\definecolor{darkmagenta}{rgb}{0.56, 0.0, 1.0}
\definecolor{softyellow}{rgb}{1.0, 0.92, 0.3} 
\definecolor{LightAquamarine}{rgb}{0.75, 1.0, 0.8} 
\definecolor{FireBrick}{RGB}{178,34,34}
\definecolor{MediumPurple}{RGB}{147,112,219}

\definecolor{uclablue}{rgb}{0.15, 0.45, 0.68}
\hypersetup{
    breaklinks,
    colorlinks=true,
    citecolor={darkmagenta},
    linkcolor={uclablue},
    urlcolor={uclablue}
}
\usepackage{wrapfig}
\usepackage{float}
\usepackage{subcaption}
\usepackage{placeins}
\usepackage{lipsum} 
\usepackage{tcolorbox}
\usepackage{amsmath}
\usepackage{amssymb}
\usepackage{utfsym}
\usepackage{fontawesome}
\usepackage{xspace}
\usepackage{enumitem}
\usepackage{multirow} 
\tcbuselibrary{breakable}
\usepackage{enumitem}
\usepackage{colortbl}
\usepackage{fancyhdr}

\usepackage{tcolorbox}
\usepackage{transparent}
\definecolor{lightgray}{gray}{0.9}
\definecolor{deepgreen}{RGB}{50,180,50}  
\definecolor{deepred}{RGB}{220,50,50}  

\definecolor{njuPurple}{RGB}{220,205,230}     
\definecolor{njuPurpleLight}{RGB}{250,245,252}   

\newtcolorbox{abstractbox}{
    colback=njuPurpleLight,   
    colframe=njuPurple,       
    boxrule=1pt,              
    arc=4mm,                  
    left=8pt,                 
    right=8pt,                
    top=8pt,                  
    bottom=8pt,               
    opacityback=0.95
}

\title{HiPO: Hybrid Policy Optimization for Dynamic Reasoning in LLMs}

\author{
Ken Deng$^{*}$,
Zizheng Zhan$^{*}$,
Wen Xiang$^{*}$,
Wenqiang Zhu$^{*}$,
Weihao Li,
Jingxuan Xu,
Tianhao Peng,
Xinping Lei,
Kun Wu,
Yifan Yao,
Haoyang Huang,
Huaixi Tang,
Kepeng Lei,
Zhiyi Lai,
Songwei Yu,
Zongxian Feng,
Zuchen Gao,
Weihao Xie,
Chenchen Zhang,
Yanan Wu,
Yuanxing Zhang,
Lecheng Huang,
Yuqun Zhang,
Jie Liu,\\
Zhaoxiang Zhang,
Haotian Zhang,
Bin Chen,
Jiaheng Liu$^{\dagger}$
\\
\vspace{4mm}
\large
\textbf{Kuaishou Technology}, \textbf{Nanjing University}
\\
\vspace{2mm}
    \texttt{dengken@kuaishou.com}, \texttt{liujiaheng@nju.edu.cn} \\
}

\begin{document}

\maketitle
\let\oldthefootnote\thefootnote

\let\thefootnote\relax\footnotetext{*~Equal Contribution. ~~$^\dagger$~Corresponding Author.}
\let\thefootnote\oldthefootnote

\input{content/0_Abstract}

\input{content/1_Introduction}

\input{content/2_relatedworks}

\input{content/3_method}
\input{content/4_expr}

\input{content/5_conclusion}

\bibliographystyle{unsrtnat}
\bibliography{references} 
\appendix
\input{content/6_appendix}
\end{document}

%% file: content/0_Abstract.tex


\begin{abstract}
\setstretch{1.2} 

Large Language Models (LLMs) increasingly rely on Chain-of-Thought (CoT) reasoning to improve accuracy on complex tasks. However, always generating lengthy reasoning traces is inefficient, leading to excessive token usage and higher inference costs. This paper introduces the Hybrid Policy Optimization (i.e., HiPO), a framework for adaptive reasoning control that enables LLMs to selectively decide when to engage in detailed reasoning (Think-on) and when to respond directly (Think-off). 
Specifically, HiPO combines a hybrid data pipeline—providing paired Think-on and Think-off responses—with a hybrid reinforcement learning reward system that balances accuracy and efficiency while avoiding over-reliance on detailed reasoning.
Experiments across mathematics and coding benchmarks demonstrate that HiPO can substantially reduce token length while maintaining or improving accuracy.
Finally, we hope HiPO~\footnote{\url{https://huggingface.co/Kwaipilot/HiPO-8B}} can be a principled approach for efficient adaptive reasoning, advancing the deployment of reasoning-oriented LLMs in real-world, resource-sensitive settings.
\end{abstract}


%% file: content/1_Introduction.tex
\section{Introduction}

Large Language Models (LLMs) have achieved unprecedented success across diverse cognitive tasks, from code generation and mathematical reasoning to scientific problem-solving. A key driver of this progress is the integration of \textbf{Chain-of-Thought (CoT)}~\citep{yao2023tree,wei2023chainofthoughtpromptingelicitsreasoning} reasoning—a paradigm where models decompose complex queries into sequential, interpretable steps to derive accurate outputs.
These approaches enhance accuracy on challenging problems but also introduce a persistent drawback: \textbf{overthinking}~\citep{kumar2025overthinkslowdownattacksreasoning,sui2025stopoverthinkingsurveyefficient,nayab2025concisethoughtsimpactoutput}. Even for trivial queries, models often generate unnecessarily long reasoning chains, leading to inflated token usage, higher latency, and reduced efficiency in interactive applications. This inefficiency creates a fundamental tension between reasoning quality and computational cost, raising the need for mechanisms that can adaptively regulate reasoning depth.

Recently, recent work has explored adaptive reasoning control to mitigate overthinking, 
and can be divided into two categories: (i) training-based adaptive reasoning, where reinforcement learning (RL)~\citep{aggarwal2025l1controllinglongreasoning,arora2025traininglanguagemodelsreason,hou2025thinkprunepruninglongchainofthought,luo2025o1prunerlengthharmonizingfinetuningo1like,shen2025dastdifficultyadaptiveslowthinkinglarge,kimiteam2025kimik15scalingreinforcement,lou2025adacotparetooptimaladaptivechainofthought} or supervised fine-tuning (SFT)~\citep{munkhdalai2024leave,ma2025cotvalvelengthcompressiblechainofthoughttuning,chen2025think23overthinkingo1like,10.1609/aaai.v39i23.34608} encourages concise reasoning through length penalties or conciseness rewards; (ii) external control, which constrains reasoning with handcrafted prompts or dynamic instructions~\citep{xu2025chaindraftthinkingfaster,Renze_2024,chen2024unlockingcapabilitiesthoughtreasoning,munkhbat2025selftrainingelicitsconcisereasoning}. While effective to some extent, these methods suffer from important limitations: coarse supervision signals, monotonic incentives that discourage deeper reasoning on difficult problems, and a lack of principled trade-offs between accuracy, latency, and token efficiency.

\begin{figure}
    \centering
    \includegraphics[width=\linewidth]{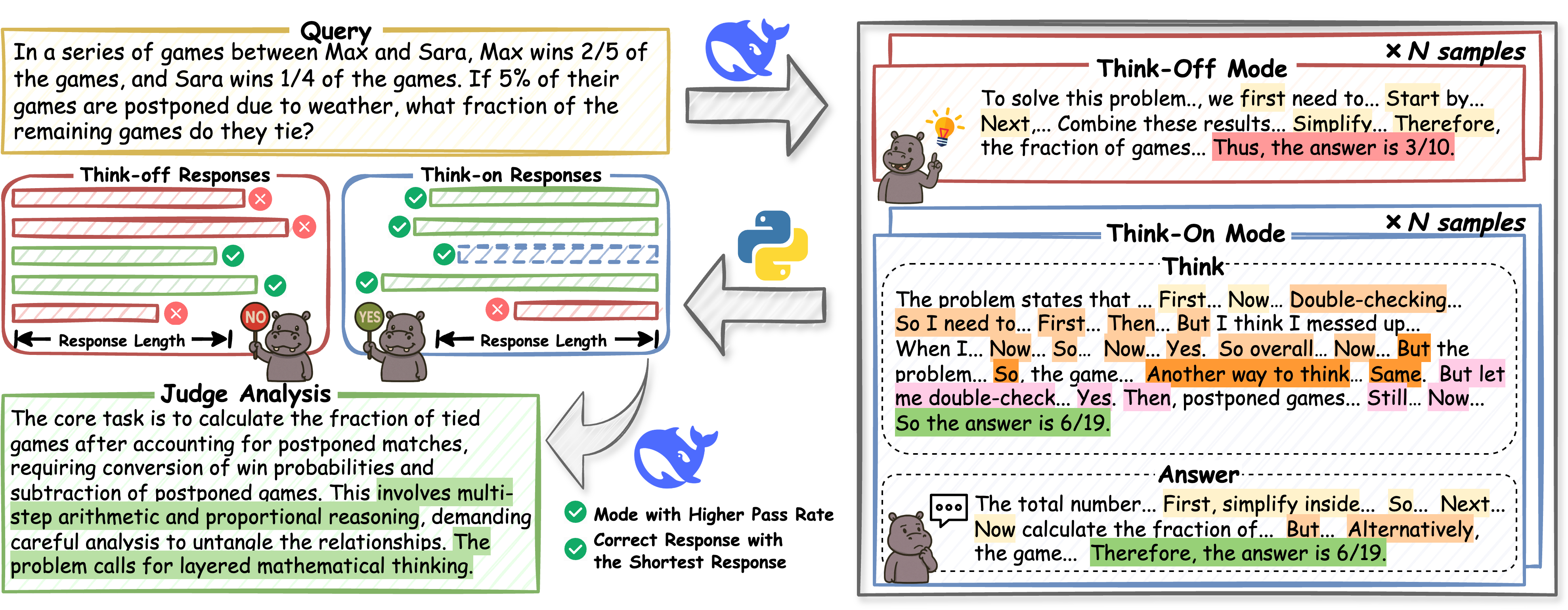}
    \caption{Framework of the hybrid data construction pipeline. }
    \label{fig:framework}
\end{figure}

To address these challenges, we introduce \textbf{HiPO} (\textbf{H}ybr\textbf{i}d \textbf{P}olicy \textbf{O}ptimization), a unified framework for adaptive reasoning in LLMs. HiPO is designed to enable models to decide when to ``think'' (i.e., \textbf{Think-on})and when to skip reasoning (i.e., \textbf{Think-off}), thereby striking a balance between correctness and efficiency. Specifically, our approach builds on two key innovations:
 (1) Hybrid Data Construction Pipeline. As shown in Figure~\ref{fig:framework}, we first collect the training data containing both Think-on and Think-off responses. Each query is automatically categorized based on its difficulty and response correctness. Then, a high-performance model (i.e., DeepSeek-V3~\citep{liu2024deepseek}) is used to produce the explicit explanations to justify its reasoning-mode decisions. Finally, for each query, the final response based on the thinking mode and the corresponding explanation construct the hybrid output.
 (2) Hybrid Reinforcement Learning Reward System. We propose a hybrid reward design that balances  Think-on and Think-off decisions. Specifically, a bias adjustment mechanism prevents the model from over-relying on verbose reasoning, while mode-aware advantage functions align reasoning-mode selection with actual performance gains. This ensures stable training and principled control over reasoning depth.

In summary, our contributions are threefold:
\begin{itemize}
    \item
We propose HiPO for adaptive LLM reasoning, which mainly includes the hybrid data construction and hybrid reinforcement learning.
    \item In the hybrid data construction pipeline, we produce logically rich Think-on and concise Think-off responses with the justification for the thinking mode. Then, for hybrid reinforcement learning, we introduce both the judge analysis and the response reward signal to enable principled control of reasoning depth.

    \item
    Experimental results on multiple datasets demonstrate that HiPO can consistently reduce redundant reasoning while improving or maintaining accuracy. 
\end{itemize}

%% file: content/2_relatedworks.tex
\section{Related Works}

\noindent\textbf{RL for LLM Reasoning.}
Recent advances in reinforcement learning (RL) have significantly enhanced LLMs' complex reasoning capabilities, moving beyond supervised fine-tuning (SFT) limitations. State-of-the-art RL algorithms demonstrate superior performance in mathematical reasoning and multi-step problem solving: GRPO ~\citep{shao2024deepseekmath} stabilizes training through intra-group relative reward comparisons; GSPO ~\citep{zheng2025groupsequencepolicyoptimization} defines sequence-level importance ratios and applies sequence-level clipping/rewarding/updates to improve efficiency and stabilize MoE training; VAPO ~\citep{yue2025vapoefficientreliablereinforcement} ensures reward consistency via value-aware optimization; PPO ~\citep{schulman2017proximalpolicyoptimizationalgorithms} constrains policy updates through clipping mechanisms; and DPO ~\citep{rafailov2024directpreferenceoptimizationlanguage} learns directly from human preferences without explicit reward modeling. 

\noindent\textbf{Adaptive Reasoning.}
Reasoning-oriented large language models—exemplified by Chain-of-Thought (CoT) ~\citep{yao2023tree,wei2023chainofthoughtpromptingelicitsreasoning} and R1-style ~\citep{deepseekai2025deepseekr1incentivizingreasoningcapability} systems—have improved complex problem solving via explicit step-by-step reasoning and self-reflection but also suffer from “overthinking” ~\citep{kumar2025overthinkslowdownattacksreasoning,sui2025stopoverthinkingsurveyefficient,nayab2025concisethoughtsimpactoutput}, where simple queries trigger redundant chains that inflate compute, latency, and token usage, hindering interactive deployment. To address this, existing work focuses on: (i) Training-based adaptive reasoning: RL to conditionally trigger CoT, length penalties and conciseness rewards~\citep{aggarwal2025l1controllinglongreasoning,arora2025traininglanguagemodelsreason,hou2025thinkprunepruninglongchainofthought,luo2025o1prunerlengthharmonizingfinetuningo1like,shen2025dastdifficultyadaptiveslowthinkinglarge,kimiteam2025kimik15scalingreinforcement,lou2025adacotparetooptimaladaptivechainofthought,zhan2025katv1kwaiautothinktechnicalreport}, and SFT~\citep{munkhdalai2024leave,ma2025cotvalvelengthcompressiblechainofthoughttuning,chen2025think23overthinkingo1like,10.1609/aaai.v39i23.34608} to prefer shorter yet correct reasoning; (ii) External control : prompt or instruction designs that limit steps or defer CoT~\citep{xu2025chaindraftthinkingfaster,Renze_2024,chen2024unlockingcapabilitiesthoughtreasoning,munkhbat2025selftrainingelicitsconcisereasoning}; (iii) Post-hoc Efficiency Optimization: pruning and restructuring chains after generation~\citep{aytes2025sketchofthoughtefficientllmreasoning,xia2025tokenskipcontrollablechainofthoughtcompression,liu2024languagemodelslearnskip,sun2024fastbestofndecodingspeculative,yang2025dynamicearlyexitreasoning}.
Despite progress, these methods still face coarse supervision, limited adaptation to hard cases due to monotonic shortening, and a lack of principled trade-offs between quality, token cost, and latency.

%% file: content/3_method.tex
\section{Method}
\label{Method}
Our HiPO framework consists of two important components: (i) a hybrid data construction pipeline that generates training data with both Think-on and Think-off responses; (ii) a hybrid reinforcement learning reward system that combines mode-specific accuracy and global average performance, along with a bias-adjustment mechanism to prevent over-reliance on the Think-on mode.

\subsection{Hybrid Data Construction Pipeline}
\label{Pipeline}
This process begins with a novel data labeling system leveraging state-of-the-art LLMs to assess each query's difficulty and domain characteristics. Queries are then classified into Think-on and Think-off categories based on their intrinsic complexity and the availability of verifiable answers.

\subsubsection{Data Source}
\begin{wrapfigure}{r}{0.4\textwidth}
    \centering
    \includegraphics[width=0.4\textwidth]{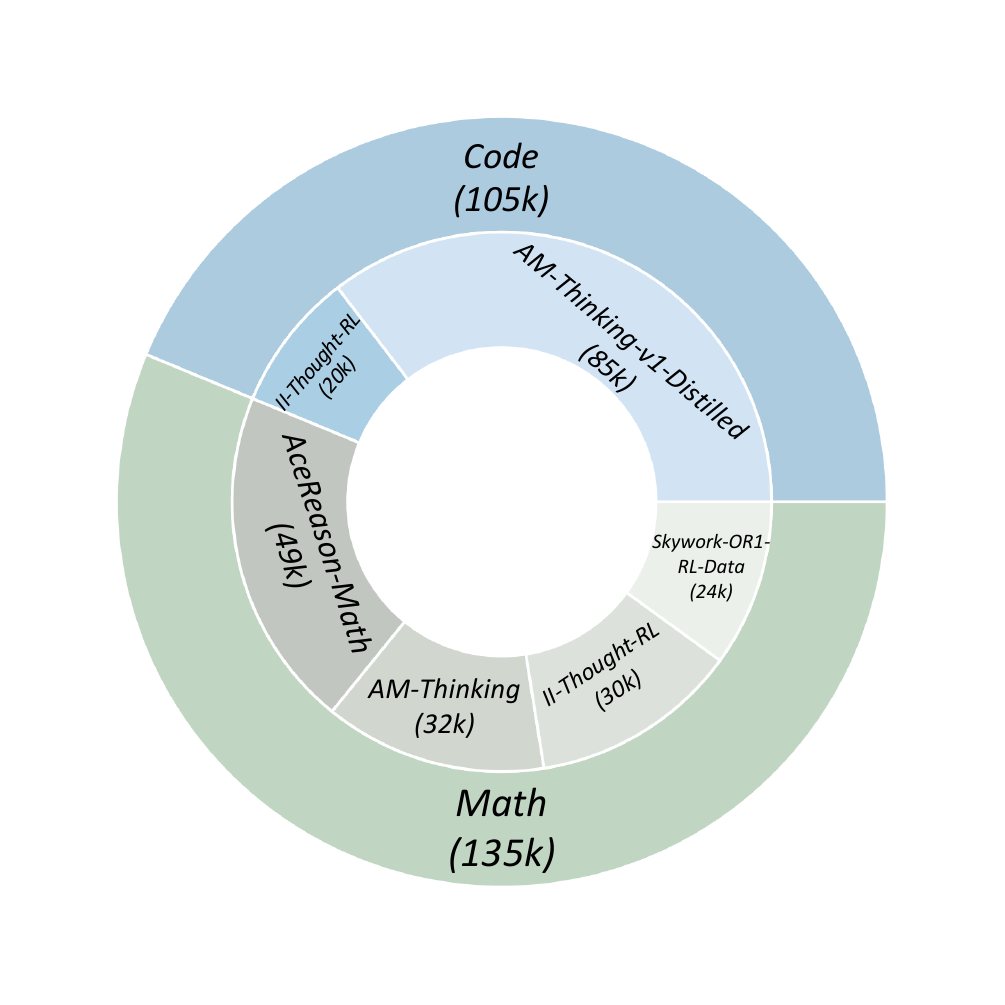}
    \caption{
    Statistics of Data Sources.
    }
    \vspace{-25pt}
    \label{fig:data_source}
\end{wrapfigure}

We construct a challenging corpus for code and mathematics by integrating diverse public and proprietary sources, as illustrated in Fig.~\ref{fig:data_source}, including AM-Thinking-v1-Distilled~\citep{tian2025not}, II-Thought-RL~\citep{2025iithought}, AceReason-Math~\citep{chen2025acereason}, and Skywork-OR1-RL-Data~\citep{skywork-or1-2025}. 
%

\subsubsection{Data Collection}

To effectively enhance the performance of HiPO, we design a structured data construction pipeline aimed at exploring and guiding the model's preference between the Think-on and Think-off reasoning modes. Our training dataset is meticulously curated to be logically rich, cross-domain, and sufficiently challenging.

We adopt a multi-stage data generation process as shown in Figure~\ref{fig:framework}. 
For each query, the pipeline samples $N$ responses under the Think-on mode and $N$ responses under the Think-off mode using a dedicated reasoning model. 
All responses are then verified for correctness, and the reasoning mode with the higher pass rate is selected as the preferred mode for that query. Let $p_{\text{on}}$ and $p_{\text{off}}$ denote the pass rates of the Think-on and Think-off modes, respectively. If the difference in pass rates satisfies $|p_{\text{on}} - p_{\text{off}}| < \delta$, where $\delta$ is a predefined threshold, the Think-off mode is selected. This tie-breaking strategy encourages the model to prefer more concise responses when deeper reasoning does not lead to a significant improvement in correctness. 
For the winning mode, 
the shortest correct response is retained as the final sample.
To expose the model to diverse reasoning scenarios and encourage adaptive behavior, we randomly assign a mode to 1\% of the queries, forcing the model to encounter diverse reasoning scenarios. This forces the model to engage with both reasoning styles in varying contexts, which is essential for learning when to switch modes dynamically during inference.
Additionally, we incorporate an auxiliary explanation signal to enhance the model’s mode alignment capabilities. For each query-response pair, we prompt DeepSeek-V3~\citep{liu2024deepseek} to generate a justification explaining why the selected mode is appropriate. This explanation provides a valuable training signal for aligning mode decisions with the underlying reasoning complexity.

\subsubsection{Data format}
The training samples follow a unified structure encompassing justification and answer generation. As shown in Table~\ref{tab:format_tokens}, this design guides the model to decide when reasoning is needed and to generate answers consistent with it. The special tokens are detailed in Table~\ref{tab:format_tokens}, ensuring a clear separation between reasoning and final response for better alignment.
\begin{table}[t]
\centering
\caption{Formatting templates (left) and special tokens with their descriptions (right).}
\begin{tabular}{c c}
\scalebox{0.59}{
\begin{tabular}{p{0.3\linewidth}|p{0.25\linewidth}}
\toprule
\textbf{Think-on Mode} & \textbf{Think-off Mode} \\
\midrule
\texttt{<judge>} & \texttt{<judge>} \\
\texttt{\textcolor{blue}{\{judge\_analysis\}}} & \texttt{\textcolor{blue}{\{judge\_analysis\}}} \\
\texttt{</judge>} & \texttt{</judge>} \\
\\
\texttt{<think\_on>} & \texttt{<think\_off>} \\
\texttt{<think>} & \texttt{<answer>} \\
\texttt{\textcolor{red}{\{thinking\_content\}}} & \texttt{\textcolor{blue}{\{response\}}} \\
\texttt{</think>} & \texttt{</answer>} \\
\\
\texttt{<answer>} & \\
\texttt{\textcolor{blue}{\{response\}}} & \\
\texttt{</answer>} & \\
\bottomrule
\end{tabular}}
&
\scalebox{0.83}{\begin{tabular}{l p{0.4\linewidth}}
\toprule
\textbf{Special Token} & \textbf{Description} \\
\midrule
\texttt{<judge>} & Analyzes input query to determine whether reasoning is required. \\
\texttt{<think\_on/off>} & Specifies whether reasoning should be activated ("on") or skipped ("off"). \\
\texttt{<think>} & Marks the beginning of reasoning in Think-on mode. \\
\texttt{<answer>} & Marks the beginning of the model’s answer. \\
\bottomrule
\end{tabular}}
\end{tabular}
\label{tab:format_tokens}
\end{table}

\subsection{Hybrid RL Reward System}
\label{RL Training}

This section details the reinforcement learning process used to teach the model how to effectively balance Think-on and Think-off reasoning modes. The approach is built on a hybrid RL reward system that guides the model's optimization.

\subsubsection{Basic Reward Formulation}
Consider a group of $N$ sampled responses, for each response $i \in \{1, \dots, N\}$, we denote its answer correctness by $\mathrm{ACC}_i \in \{0,1\}$, its format correctness by $\mathrm{FORMAT}_i \in \{0,1\}$, its basic reward by $r_i \text{=} \mathrm{ACC}_i + 0.2 \cdot \mathrm{FORMAT}_i \in \mathbb{R}$, and its reasoning mode by $M_i \in \{\text{on}, \text{off}\}$, where $M_i \text{=} \text{on}$ indicates the Think-on mode and $M_i \text{=} \text{off}$ indicates the Think-off mode.



\subsubsection{Bias Adjustment Mechanism}
\label{Bias Adjustment Mechanism}
A potential risk of the hybrid reward design is that the model may overfit to the more accurate Think-on mode, favoring deep reasoning even when it is unnecessary. This tendency can reduce response efficiency and hinder the intended flexibility in reasoning behavior. 
To mitigate this issue, we introduce a bias adjustment mechanism that dynamically regularizes the contribution of mode-specific accuracies. 

Let $\text{mean}(\mathbf{r}_{\text{on}})\text{=}\frac{1}{N_{\text{on}}} \sum_{i: M_i = \text{on}} r_i$ denote the average reward of responses generated under the Think-on mode, and let $\text{mean}(\mathbf{r}_{\text{off}})$ denote the corresponding average reward for the Think-off mode.
Based on this, we define a bias term for the Think-off mode as a fraction of the Think-on average reward:
$\mathrm{bias}_{\text{off}} \text{=} \omega \cdot \text{mean}(\mathbf{r}_{\text{on}})$, where $\omega$ controls the ratio.
The adjustment is applied only when the performance of the Think-off mode does not exceed that of the Think-on mode, but the difference between the two remains within the bias threshold. Formally, the adjustment mechanism is as follows:
\begin{equation}
\text{mean}(\mathbf{r}_{\text{off}}) =
\begin{cases}
\text{mean}(\mathbf{r}_{\text{off}}) + \mathrm{bias}_{\text{off}}, &   
0 \leq \text{mean}(\mathbf{r}_{\text{on}}) - \text{mean}(\mathbf{r}_{\text{off}}) \leq \mathrm{bias}_{\text{off}}, \\[6pt]
\text{mean}(\mathbf{r}_{\text{off}}), & \text{otherwise}.
\end{cases}
\end{equation}

This mechanism prevents the model from gaining an unfair advantage by overfitting to the more verbose but more accurate Think-on mode. Moreover, it ensures that the adjusted accuracies remain faithful to the true relative performance between reasoning modes, thereby improving training stability and preserving the intended balance between depth and efficiency.

\subsubsection{Supervision RL with HiPO}
\label{Supervision RL with HiPO}
The final advantage function is formulated as a hybrid signal that integrates both judge analysis and model response. Each response $i$ receives two distinct scalar advantage, including \emph{judge advantage} based on the quality of the mode justification, and \emph{answer advantage} based on correctness and format.

The judge advantage $\mathrm{A}^{\text{judge}}_i$ captures the broader decision-level utility of selecting a particular mode. The first term, $\text{mean}(\mathbf{r}_{M_i}) - \text{mean}(\mathbf{r})$, quantifies the global advantage of the chosen mode over the full group average, guiding the model toward choosing modes that lead to higher expected rewards. The second term, $\gamma \cdot (r_i - \text{mean}(\mathbf{r}))$, ensures that the justification content is also responsible for the quality of the response under that mode, thereby aligning the explanation with actual performance. The use of a global normalization factor $\text{std}(\mathbf{r})$ stabilizes the reward signal across groups. 
The judge advantage function for response $i$ is then given by:
\begin{equation}
\mathrm{A}^{\text{judge}}_i =
\begin{cases}
\frac{(\text{mean}(\mathbf{r}_{\text{on}}) - \text{mean}(\mathbf{r})) + \gamma \cdot (r_i - \text{mean}(\mathbf{r}))}{\text{std}(\mathbf{r})}, & \text{if } M_i=\text{on}, \\[6pt]
\frac{(\text{mean}(\mathbf{r}_{\text{off}}) - \text{mean}(\mathbf{r})) + \gamma \cdot (r_i - \text{mean}(\mathbf{r}))}{\text{std}(\mathbf{r})}, & \text{if } M_i=\text{off}.
\end{cases}
\end{equation}

In contrast to the judge advantage function, the advantage $A^{\text{answer}}_i$ is computed within the context of the selected reasoning mode. Since the mode $M_i$ has already been determined prior to response generation, it is natural to assess the response quality relative to other responses within the same mode. This local normalization using mode-specific mean and standard deviation focuses the learning signal on intra-mode variance, encouraging the model to improve response quality without conflating mode preference. For response $i$, the answer advantage is defined as:
\begin{equation}
A^{\text{answer}}_i =
\begin{cases}
\frac{r_i - \text{mean}(\mathbf{r}_{\text{on}})}{\text{std}(\mathbf{r}_{\text{on}})}, & \text{if } M_i=\text{on}, \\[6pt]
\frac{r_i - \text{mean}(\mathbf{r}_{\text{off}})}{\text{std}(\mathbf{r}_{\text{off}})}, & \text{if } M_i=\text{off}.
\end{cases}
\end{equation}

To assign token-level reward for training with reinforcement learning, we define the final reward for each token $t$ in sample $i$ as follows:
\begin{equation}
\mathrm{A}_{i,t} =
\begin{cases}
\mathrm{A}^{\text{answer}}_i, & \text{if token } t \in \mathcal{T}^{\text{answer}}, \\
\mathrm{A}^{\text{judge}}_i, & \text{if token } t \in \mathcal{T}^{\text{judge}}.
\end{cases}
\label{eq_adv}
\end{equation}
where $\mathcal{T}^{\text{judge}}$ and $\mathcal{T}^{\text{answer}}$ denote the token index sets corresponding to the judge segment and the answer segment, respectively, within each response.


Given a query $q$, HiPO generates a collection of candidate outputs $\{o_i\}_{i=1}^{G}$ from the old policy $\pi_{\theta_{\text{old}}}$. For each output $o_i$, let $\mathcal{T}_i$ denote the set of token positions in response $i$, i.e., $\mathcal{T}_i = \mathcal{T}^{\text{judge}} \cup \mathcal{T}^{\text{answer}}$. We define the per-token probability ratio as
$\rho_{i,t} 
    = \frac{\pi_{\theta}\!\left(y_{i,t} \,\middle|\, h_{i,t}\right)}{\pi_{\theta_{\text{old}}}\!\left(y_{i,t} \,\middle|\, h_{i,t}\right)}$,
where $y_{i,t}$ is the $t$-th generated token in $o_i$ and $h_{i,t}$ is its conditioning context.
The policy $\pi_{\theta}$ is optimized by maximizing the following token-level objective:
\begin{equation}
\begin{aligned}
    \mathcal{J}(\theta)
    &= \mathbb{E}\Big[ q \sim P(Q), \{o_i\}_{i=1}^{G} \sim \pi_{\theta_{\text{old}}}(\cdot\,|\,q) \Big] \\
    &\quad \cdot \frac{1}{G} \sum_{i=1}^{G} \frac{1}{\lvert \mathcal{T}_i \rvert} \sum_{t \in \mathcal{T}_i}
    \Big(
        \min\big( \rho_{i,t} \, A_{i,t},\; \text{clip}(\rho_{i,t},\,1-\epsilon,\,1+\epsilon) \, A_{i,t} \big)
        \\
        &\qquad - \beta \, \mathbb{D}_{KL}\!\big( \pi_{\theta}(\cdot\,|\,h_{i,t}) \,\big\|\, \pi_{\text{ref}}(\cdot\,|\,h_{i,t}) \big)
    \Big) .
\end{aligned}
\end{equation}
Here $A_{i,t}$ is the token-level advantage defined in Eq. (\ref{eq_adv}) via segment-wise assignment, and $\mathbb{D}_{KL}$ is the token-level KL between the current policy and the reference policy at context $h_{i,t}$.

\subsection{Training Paradigm}
Our HiPO framework adopts a two-stage training paradigm, consisting of a \textbf{cold-start} stage and a \textbf{RL} stage. 
In the code-start stage, the model is initialized with high-quality, hybrid training data that contains both Think-on and Think-off responses. 
This stage enables the model to acquire fundamental reasoning and answering capabilities, while establishing an initial balance between analytical reasoning and concise responses. 
In the RL stage, the model is further optimized using our hybrid reward system, which integrates mode-specific accuracy and global average performance. 
Together, these two stages ensure that HiPO achieves both strong factual accuracy and robust reasoning ability across diverse domains.

%% file: content/4_expr.tex
\section{Experiments}

\subsection{Experimental setup}\label{sec:exp_setup}

\noindent\textbf{Implementation details.}
Since the Qwen3 model can freely switch between inference modes, we chose it for our experiment. However, when the training data is insufficient, training the Qwen3 model can easily lead to a decline in performance on the test set (details can be found in the appendix \ref{qwen3_decline}). 
To address this, we conducted Cold-Start tuning to stabilize its performance with relatively large datasets.
For the Cold-Start stage, we use the ``AM-Thinking-v1-Distilled'', ``AceReason-Math'', ``AM-Thinking'', ``II-Thought-RL(math)'' dataset for training. The parameters are set as: maximum learning rate is 8e-5, minimum learning rate is 8e-6
and batch size is 512. For the RL stage, we use the ``II-Thought-RL(code)'', ``Skywork-OR1-RL-Data'' dataset for training. The parameters are set as: batch size = 16, maximum response length = 32k, $N$ = 16, $\omega$ = 0.01, and $\gamma = 0.3$.

\noindent\textbf{Baselines.}
To demonstrate the effect of mitigating overthinking, we designed the following baselines for comparison.
(1) \textbf{Cold-Start}: We perform Cold-Star on the model using the data construction method described in Section~\ref{Pipeline}.
(2) \textbf{Cold-Start (On)}: We apply the same Cold-Star procedure as in Section~\ref{Pipeline}, but only include the data collected under the Think-on mode.
(3) \textbf{Cold-Start (On) + GRPO}: We further train the \textbf{Cold-Start (On)} model using the GRPO algorithm.
(4) \textbf{Cold-Start + GRPO}: We further train the \textbf{Cold-Start} model with the GRPO algorithm.
(5) \textbf{HiPO}: We train the model following our HiPO.
(6) \textbf{AdaptThink}: We reproduced the code provided in \citep{zhang2025adaptthinkreasoningmodelslearn}.
(7) \textbf{AutoThink}: We reproduced the code provided in \citep{tu2025learningthinkshapingadaptive}.


\noindent\textbf{Evaluation benchmarks.}
We conducted tests on AIME2024, AIME2025, HumanEval~\citep{chen2021evaluatinglargelanguagemodels}, LiveCodeBench V6~\citep{jain2024livecodebenchholisticcontaminationfree}, MBPP~\citep{austin2021programsynthesislargelanguage},
MATH-500~\citep{lightman2023letsverifystepstep}, and GPQA-Diamond~\citep{Rein2023GPQAAG}.

\subsection{Main Results}

\input{Tables/main}

In Table \ref{tab:main}, we observe that training the model solely on Think-on data leads the model to engage in reasoning for problems of any difficulty. We use this baseline as a typical example of "overthinking" for comparison. After applying GRPO to the Cold-Start (on) model, there is a significant improvement in accuracy, with an average accuracy increase of 3.1\%. However, this does not reduce the token length and thinking rate of the model. On the contrary, to achieve higher accuracy, the token length output by the model on simpler datasets increases significantly. When training the model on a dataset containing both Think-on and Think-off data, the accuracy of the resulting Cold-Start model improves by 4.0\% compared to the Cold-Start(on) model, while the token length and thinking rate decrease by 10.8\% and 22\%, respectively. After applying the GRPO algorithm to the Cold-Start model, there is no significant change in performance. However, when applying our method HiPO to train the Cold-Start model, the accuracy improves by 6.2\%, while the token length and thinking rate decrease dramatically by 30\% and 39\%, respectively. Moreover, experimental results show that our HiPO outperforms existing methods on both efficiency and accuracy. 

\subsection{Ablation Study}

\begin{figure}
    \centering
    \includegraphics[width=1.0\linewidth]{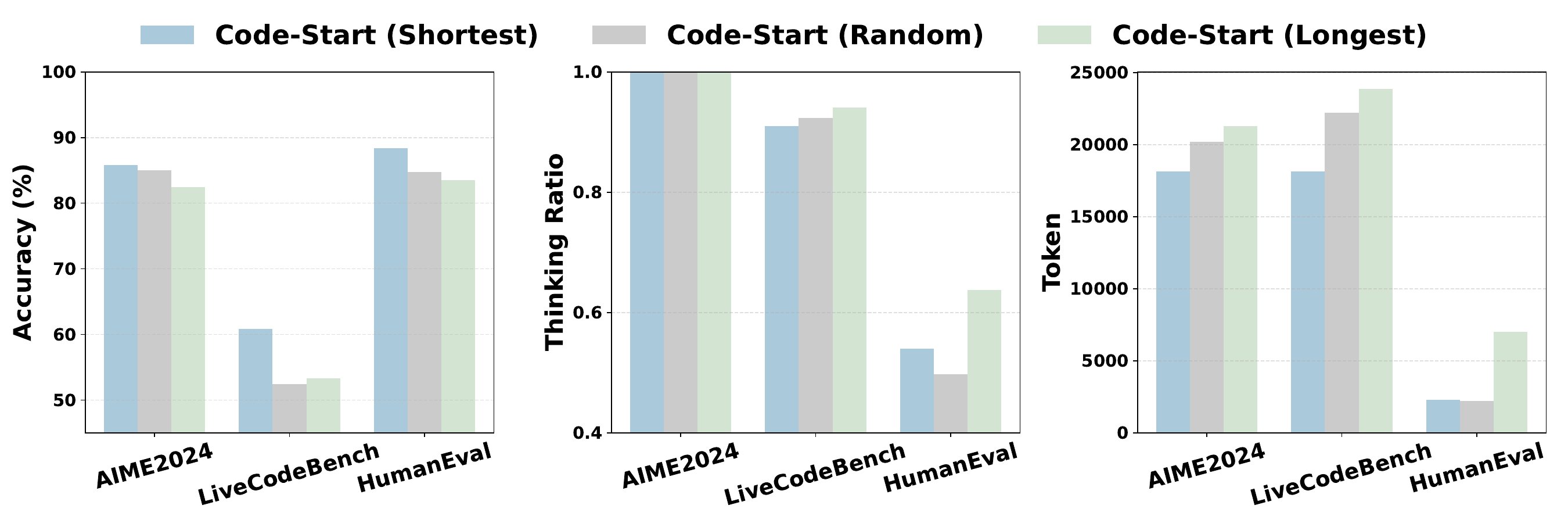}
    \caption{Performance of different response selection strategies.}
    \label{fig:SFT}
\end{figure}

\noindent\textbf{Effect of selecting the shortest response.}
\label{Ablation Study on Cold-start Stage}
In the data construction pipeline, we select the shortest response (Cold-Start (Shortest)) as the final sample.
To analyze the effect of this strategy,
we additionally propose two variants called (\textbf{Cold-Start (Longest)} and \textbf{Cold-Start (Random)}) by selecting the longest responses and randomly selecting the responses,
respectively.
In Figure~\ref{fig:SFT},
Cold-Start (Shortest) shows an improvement in accuracy compared to both Cold-Start (Longest) and Cold-Start (Random), with a decrease in both the Thinking ratio and Token length.
Therefore, 
we adopt this Cold-Start (Shortest) strategy for the Cold-Start stage.

\noindent\textbf{Effect of design strategies for $\mathrm{A}^{\text{judge}}_i$ and $\mathrm{A}^{\text{answer}}_i$.}
\label{Ablation Study on hipo}
In the reinforcement learning stage, first, we utilize the term $\text{mean}(\mathbf{r}_{M_i}) - \text{mean}(\mathbf{r})$ to quantify the \textbf{global advantage} of the chosen mode over the full group average.  Second, the \textbf{local normalization} based on the mode-specific mean and standard deviation is used for $\mathrm{A}^{\text{answer}}_i$.
To demonstrate the effect of these strategies, as shown in Table~\ref{a_variants}, we design two variants (i.e., HiPO (w/o global adv) and HiPO (w/o local norm)). For HiPO (w/o global adv), we directly remove the global advantage for  $\mathrm{A}^{\text{judge}}_i$. For HiPO (w/o local norm), we just use the global normalization across the responses in a group.
In Table~\ref{a_variants}, we observe that  HiPO achieves significant improvements in performance and efficiency when compared to these two variants.

\input{Tables/RL_Ablation}

\begin{figure}[t]
    \centering
    \includegraphics[width=1.0\linewidth]{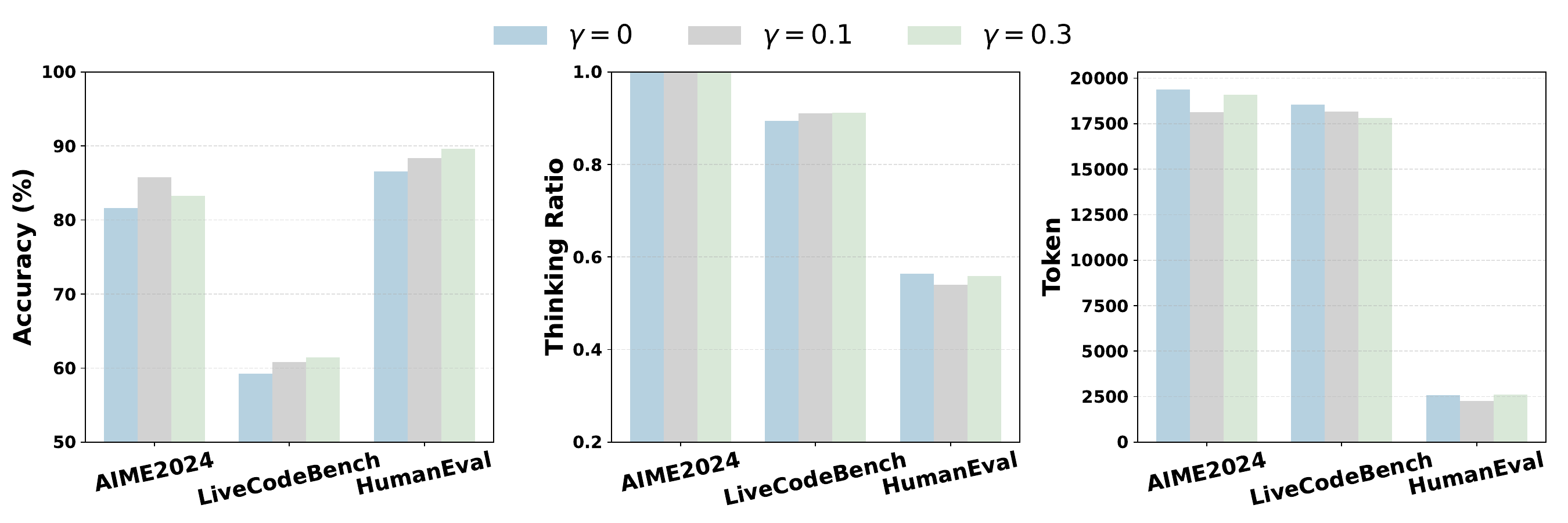}
    \caption{Performance of different $\gamma$ values.}

    \label{fig:gamma}
\end{figure}

\begin{figure}[t]
    \centering
    \includegraphics[width=1.0\linewidth]{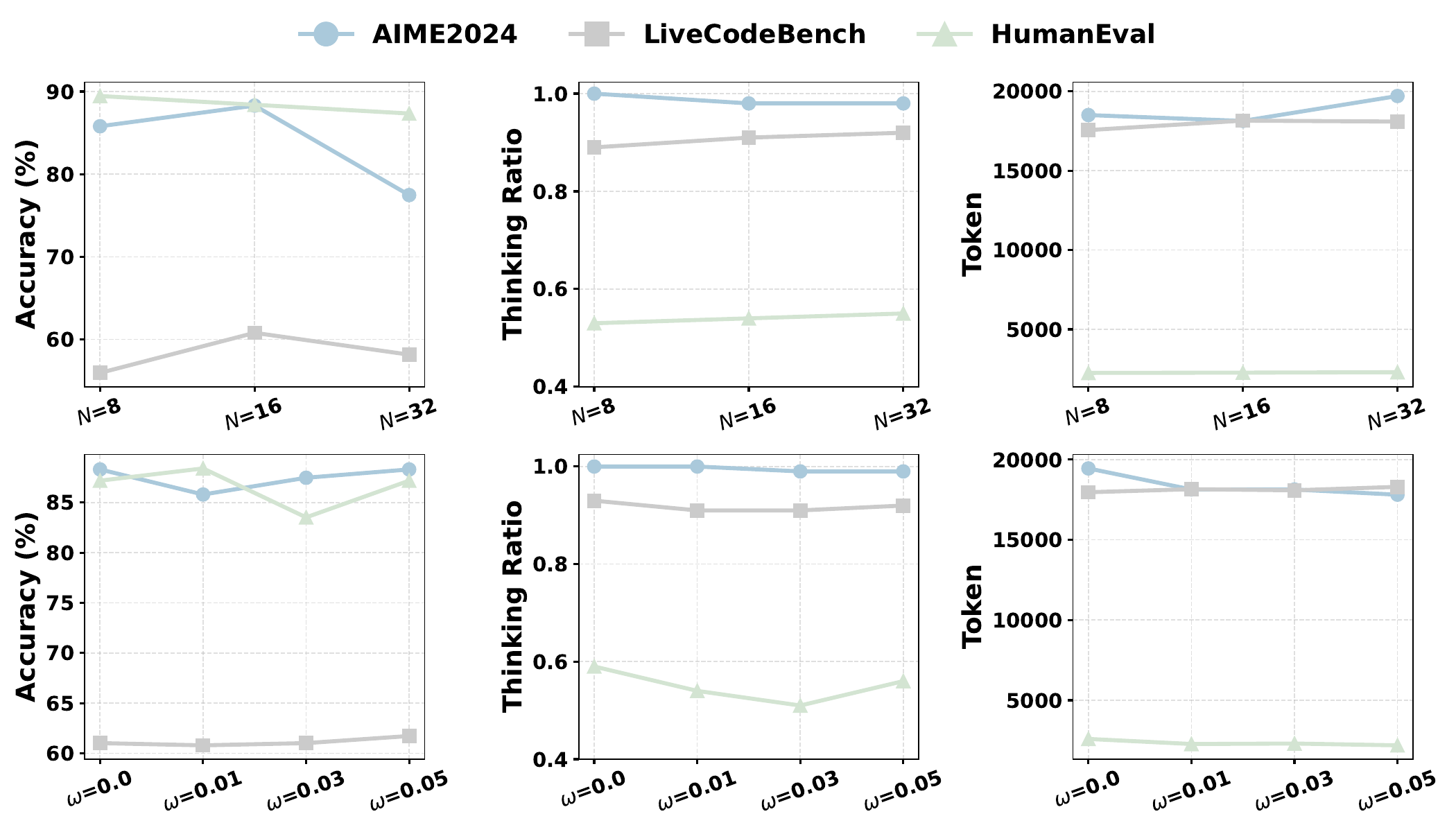}

        \caption{Performance of different rollout numbers and $\omega$ values.}

    \label{fig:omega_and_N}
\end{figure}

\textbf{Effect of different $\gamma$ values.} Figure \ref{fig:gamma} shows that, when the value of $\gamma$ is set to 0.00, the reward for the judge token lacks information about the current response, resulting in lower model accuracy and higher token length. On the other hand, when $\gamma$ is set too high, the scales of the two terms (\textit{mean}($\mathbf{r}_{\text{off}}$) - \textit{mean}($\mathbf{r}$)) and $(r_i - \textit{mean}(\mathbf{r}))$ become imbalanced, which leads to a decrease in model accuracy and an increase in token length.

\textbf{Effect of different rollout numbers.} Table \ref{fig:omega_and_N} shows that, when the rollout number $N$ is set to 16, the model achieves better average performance, shorter token length, and lower think rate. We attribute this to the fact that this configuration provides sufficient data to explore diverse possibilities while avoiding excessive samples with redundant reasoning that dilute the training signal. As a result, the model focuses more on learning from higher-quality samples, leading to a more concise strategy with improved accuracy, reduced token length, and lower think rate.

\textbf{Effect of different $\omega$ values.} Table \ref{fig:omega_and_N} shows that, setting $\omega$ to 0.01 provides a balanced trade-off between performance and efficiency.  This configuration mitigates the overly conservative behavior seen at 0.0 while avoiding the overly aggressive behavior at higher settings, ultimately achieving the largest efficiency gains with minimal performance loss.

\subsection{Further Analysis}
\begin{figure}[t]
    \centering
    \subfigure[Training]{
        \includegraphics[width=0.45\linewidth]{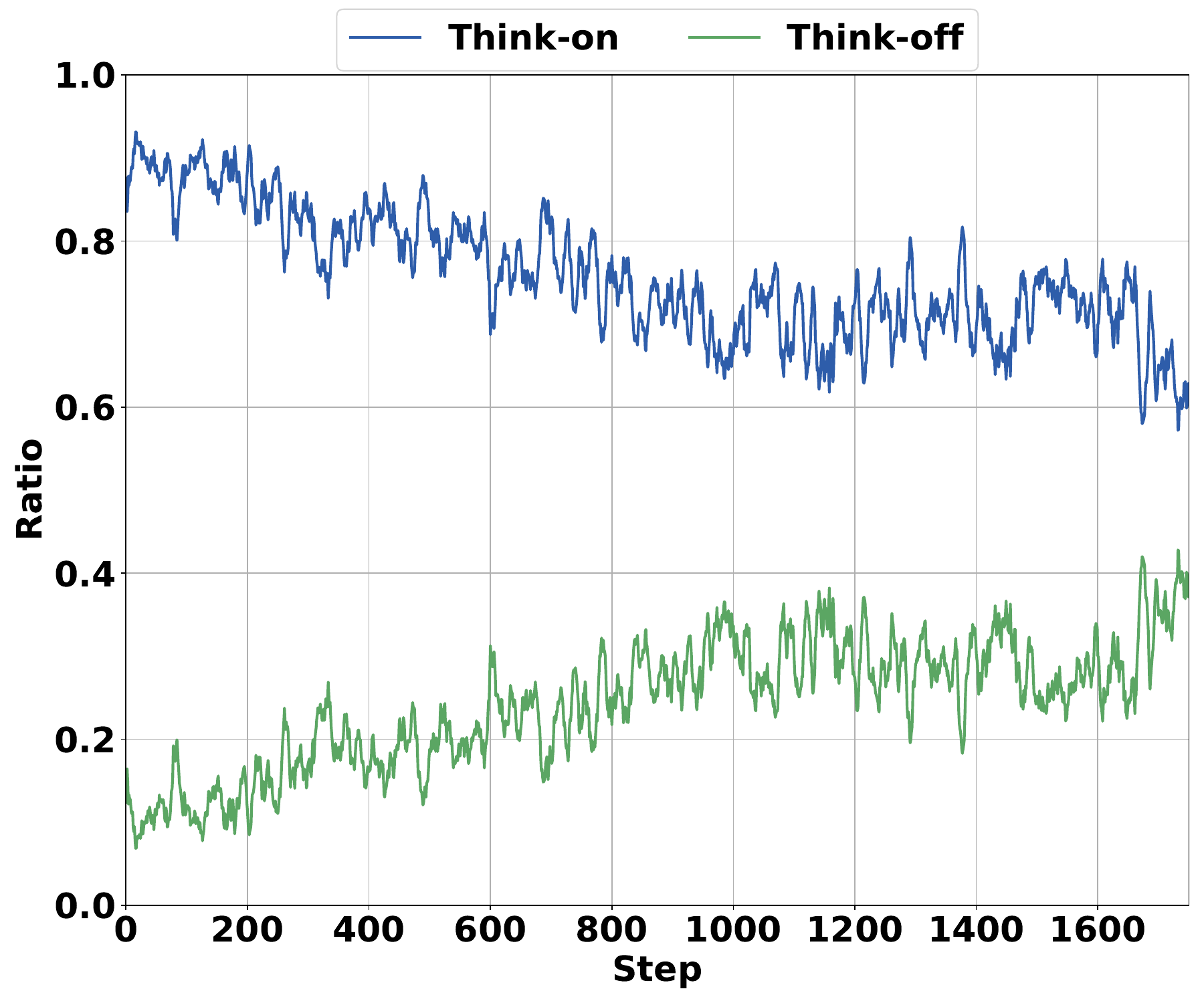}
        \label{fig:Training Ratio}
    }
    \subfigure[Testing]{
        \includegraphics[width=0.45\linewidth]{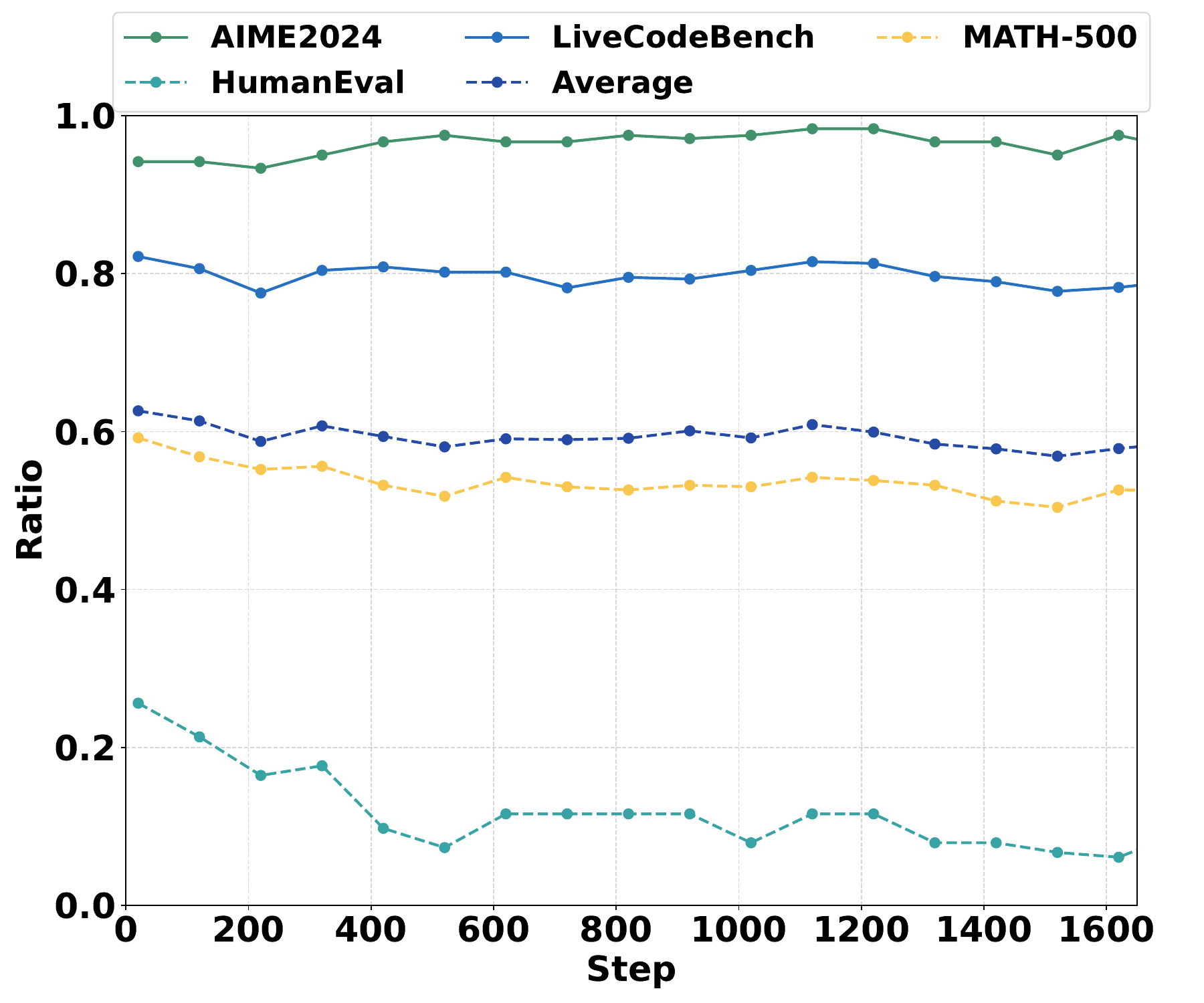}
        \label{fig:Testing Ratio}
    }
    \caption{(a) Think-on and Think-off ratio in training. (b) Think-on ratio of different datasets.}
    \label{fig:ratio dynamics}
\label{gamma-1}
\end{figure}
\begin{figure}[t]
    \centering
    \subfigure[Training]{
        \includegraphics[width=0.45\linewidth]{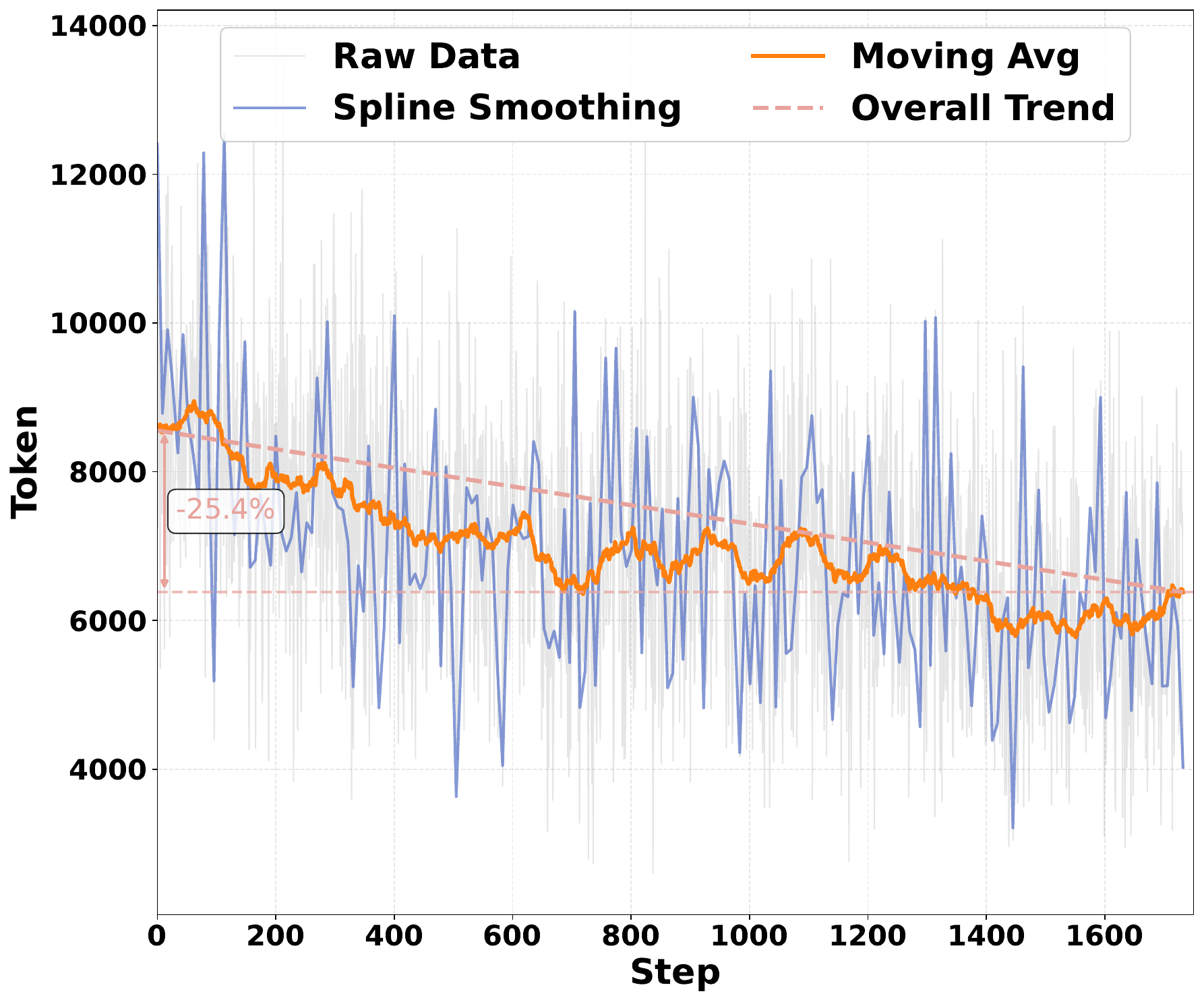}
        \label{fig:Training Token}
    }
    \subfigure[Testing]{
        \includegraphics[width=0.45\linewidth]{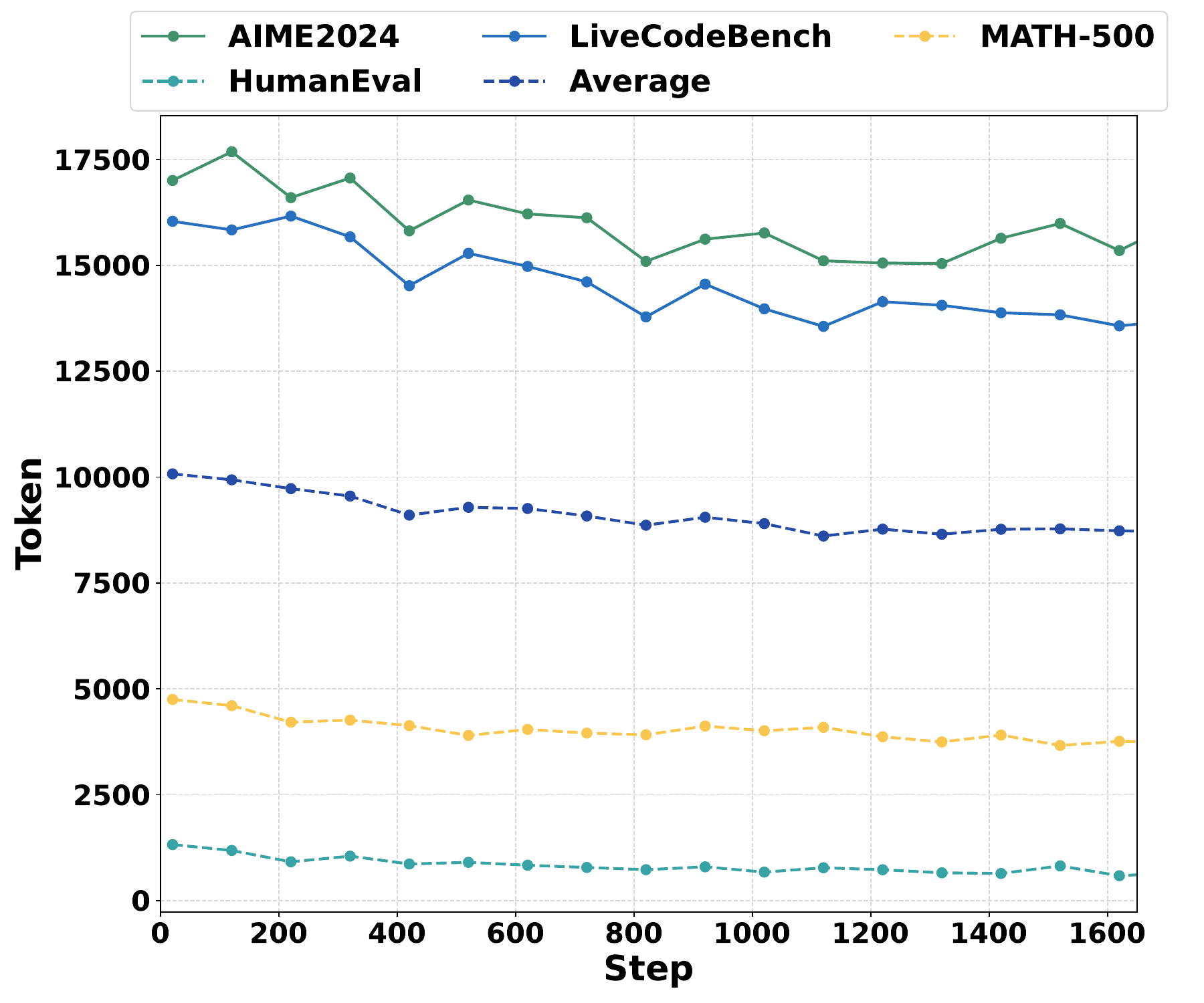}
        \label{fig:Testing Token}
    }
    \caption{(a) Average token usage in RL training. (b) Token usage of different datasets.}
    \label{fig:Token dynamics}
\label{gamma-2}
\end{figure}


\begin{table}[!htp]
\centering
\scriptsize
\resizebox{\linewidth}{!}{
    \begin{tabular}{l|ccc|ccc|ccc|ccc}
    \toprule
      & \multicolumn{3}{c|}{\cellcolor{gray!10}\textbf{AIME24}} & \multicolumn{3}{c|}{\cellcolor{gray!10}\textbf{LiveCodeBench}} & \multicolumn{3}{c|}{\cellcolor{gray!10}\textbf{HumanEval}} & \multicolumn{3}{c}{\cellcolor{gray!10}\textbf{MBPP}}  \\
      \cmidrule(lr){2-4} \cmidrule(lr){5-7} \cmidrule(lr){8-10} \cmidrule(lr){11-13}
    \multirow{-2}{*}{\textbf{Method}} & \textbf{Acc} & \textbf{Length} & $\textbf{Ratio}_\textbf{T}$ & \textbf{Acc} & \textbf{Length} & $\textbf{Ratio}_\textbf{T}$ & \textbf{Acc} & \textbf{Length} & $\textbf{Ratio}_\textbf{T}$ & \textbf{Acc} & \textbf{Length} & $\textbf{Ratio}_\textbf{T}$  \\ \midrule
    \multicolumn{13}{l}{\textbf{Qwen3-1.7B}} \\ \midrule
    Cold-Start (On) & 63.3 & 24214 & 1.00 & 33.7 & 25616 & 1.00 & 77.4 & 4172 & 1.00 & 54.6 & 8587 & 1.00 \\

Cold-Start & 65.0\textcolor{deepgreen}{$\scriptstyle\uparrow{\scriptstyle 2.7\%}$} & 21039\textcolor{deepgreen}{$\scriptstyle\downarrow{\scriptstyle 13.1\%}$} & 1.00\textcolor{uclablue}{$\scriptstyle-{\scriptstyle 0.0\%}$} & 37.4\textcolor{deepgreen}{$\scriptstyle\uparrow{\scriptstyle 11.0\%}$} & 21364\textcolor{deepgreen}{$\scriptstyle\downarrow{\scriptstyle 16.6\%}$} & 0.98\textcolor{deepgreen}{$\scriptstyle\downarrow{\scriptstyle 2.0\%}$} & 81.7\textcolor{deepgreen}{$\scriptstyle\uparrow{\scriptstyle 5.2\%}$} & 3084\textcolor{deepgreen}{$\scriptstyle\downarrow{\scriptstyle 26.1\%}$} & 0.39\textcolor{deepgreen}{$\scriptstyle\downarrow{\scriptstyle 61.0\%}$} & 54.4\textcolor{deepred}{$\scriptstyle\downarrow{\scriptstyle 0.4\%}$} & 6398\textcolor{deepgreen}{$\scriptstyle\downarrow{\scriptstyle 25.5\%}$} & 0.64\textcolor{deepgreen}{$\scriptstyle\downarrow{\scriptstyle 36.0\%}$} \\
  
    HiPO& \underline{68.3}\textcolor{deepgreen}{$\scriptstyle\uparrow{\scriptstyle 7.9\%}$} & \underline{17614}\textcolor{deepgreen}{$\scriptstyle\downarrow{\scriptstyle 27.3\%}$} & \underline{0.98}\textcolor{deepgreen}{$\scriptstyle\downarrow{\scriptstyle 6\%}$} & \underline{44.3}\textcolor{deepgreen}{$\scriptstyle\uparrow{\scriptstyle 31.4\%}$} & \underline{19358}\textcolor{deepgreen}{$\scriptstyle\downarrow{\scriptstyle 24.4\%}$} & \underline{0.92}\textcolor{deepgreen}{$\scriptstyle\downarrow{\scriptstyle 8.0\%}$} & \underline{86.0}\textcolor{deepgreen}{$\scriptstyle\uparrow{\scriptstyle 11.1\%}$} & \underline{1973}\textcolor{deepgreen}{$\scriptstyle\downarrow{\scriptstyle 52.7\%}$} & \underline{0.28}\textcolor{deepgreen}{$\scriptstyle\downarrow{\scriptstyle 62.0\%}$} & \underline{62.8}\textcolor{deepgreen}{$\scriptstyle\uparrow{\scriptstyle 15.0\%}$} & \underline{4330}\textcolor{deepgreen}{$\scriptstyle\downarrow{\scriptstyle 49.6\%}$} & \underline{0.47}\textcolor{deepgreen}{$\scriptstyle\downarrow{\scriptstyle 53.0\%}$}\\ 
    \midrule
        \multicolumn{13}{l}{\textbf{Qwen3-32B}} \\ \midrule
            Cold-Start (On) & 81.7 & 19551 & 1.00 & 65.4 & 17885 & 1.00 & 87.8 & 4298 & 1.00 & 76.2 & 4753 & 1.00 \\

Cold-Start & 85.0\textcolor{deepgreen}{$\scriptstyle\uparrow{\scriptstyle 4.3\%}$} & 16542\textcolor{deepgreen}{$\scriptstyle\downarrow{\scriptstyle 15.4\%}$} & 1.00\textcolor{uclablue}{$\scriptstyle{\scriptstyle -0.0\%}$} & 65.9\textcolor{deepgreen}{$\scriptstyle\uparrow{\scriptstyle 0.8\%}$} & 14935\textcolor{deepgreen}{$\scriptstyle\downarrow{\scriptstyle 16.5\%}$} & 0.87\textcolor{deepgreen}{$\scriptstyle\downarrow{\scriptstyle 13.0\%}$} & 92.1\textcolor{deepgreen}{$\scriptstyle\uparrow{\scriptstyle 4.9\%}$} & 2785\textcolor{deepgreen}{$\scriptstyle\downarrow{\scriptstyle 35.2\%}$} & 0.47\textcolor{deepgreen}{$\scriptstyle\downarrow{\scriptstyle 53.0\%}$} & 78.4\textcolor{deepgreen}{$\scriptstyle\uparrow{\scriptstyle 2.2\%}$} & 3991\textcolor{deepgreen}{$\scriptstyle\downarrow{\scriptstyle 16.0\%}$} & 0.51\textcolor{deepgreen}{$\scriptstyle\downarrow{\scriptstyle 49.0\%}$} \\
 
    HiPO& \underline{88.3}\textcolor{deepgreen}{$\scriptstyle\uparrow{\scriptstyle 8.1\%}$} & \underline{14873}\textcolor{deepgreen}{$\scriptstyle\downarrow{\scriptstyle 23.9\%}$} & \underline{0.98}\textcolor{deepgreen}{$\scriptstyle\downarrow{\scriptstyle 2.0\%}$} & \underline{68.5}\textcolor{deepgreen}{$\scriptstyle\uparrow{\scriptstyle 4.5\%}$} & \underline{12721}\textcolor{deepgreen}{$\scriptstyle\downarrow{\scriptstyle 28.9\%}$} & \underline{0.82}\textcolor{deepgreen}{$\scriptstyle\downarrow{\scriptstyle 18.0\%}$} & \underline{92.7}\textcolor{deepgreen}{$\scriptstyle\uparrow{\scriptstyle 5.6\%}$} & \underline{824}\textcolor{deepgreen}{$\scriptstyle\downarrow{\scriptstyle 80.8\%}$} & \underline{0.18}\textcolor{deepgreen}{$\scriptstyle\downarrow{\scriptstyle 82.0\%}$} & \underline{84.4}\textcolor{deepgreen}{$\scriptstyle\uparrow{\scriptstyle 10.8\%}$} & \underline{2070}\textcolor{deepgreen}{$\scriptstyle\downarrow{\scriptstyle 56.4\%}$} & \underline{0.24}\textcolor{deepgreen}{$\scriptstyle\downarrow{\scriptstyle 76.0\%}$}\\ 
    \midrule
    \end{tabular}
}
\caption{Performance of HiPO on more models.}
\label{tab:moremodels}
\end{table}

We analyze two key dimensions: (i) reasoning-mode activation (\texttt{<think\_on>} vs. \texttt{<think\_off>}) and (ii) token efficiency across RL training steps and benchmark tasks.
Specifically, during the training and evaluation processes, we track how the model’s decision-making evolves by monitoring the frequency of reasoning-mode activations and the corresponding output length.

\paragraph{Think-on vs. Think-off Dynamics During Training and Inference}
We logged the frequency of \texttt{<think\_on>} and \texttt{<think\_off>} activations at each step. As shown in Figure~\ref{fig:Training Ratio}, HiPO not only improves final accuracy but also sharpens the model’s gating behavior, allowing it to skip unnecessary reasoning. Specifically, the gap between \texttt{<think\_on>} and \texttt{<think\_off>} activations decreases from 89.5\% at the beginning of training to 53.1\% by the end. In Figure \ref{fig:Testing Ratio} shows the proportion of Think-on activations across different datasets during inference. Reasoning-intensive tasks, including AIME2024, and LiveCodeBench, consistently demonstrate high Think-on activation rates (>70\%) throughout training. Conversely, tasks that require less explicit reasoning, such as HumanEval --- exhibit a clear downward trend in Think-on activation as training progresses.

\paragraph{Token Count Dynamics During Training and Inference}
During RL training, the average token count shows a consistent downward trend in Figure~\ref{fig:Training Token}, which indicates that the model gradually learns to produce more concise responses and highlight the HiPO reward design in encouraging efficient token usage
Besides, Figure \ref{fig:Testing Token} shows the corresponding dynamics in average token counts per generated response during inference, and we also observe consistent token reduction in training.

\paragraph{Generalization on More Models} In Table~\ref{tab:moremodels}, we report the performance of HiPO on Qwen3-1.7B and Qwen3-32B, which shows consistent improvements on both accuracy and efficiency.


%% file: Tables/main.tex
\begin{table}[H]
\centering
\scriptsize
\resizebox{\linewidth}{!}{
    \setlength{\tabcolsep}{1pt}
    \renewcommand{\arraystretch}{1.5}
    \begin{tabular}{l|ccc|ccc|ccc|ccc}
    \toprule
      & \multicolumn{3}{c|}{\cellcolor{gray!10}\textbf{AIME2024}} & \multicolumn{3}{c|}{\cellcolor{gray!10}\textbf{AIME2025}} & \multicolumn{3}{c|}{\cellcolor{gray!10}\textbf{LiveCodeBench}} & \multicolumn{3}{c}{\cellcolor{gray!10}\textbf{HumanEval}}  \\
      \cmidrule(lr){2-4} \cmidrule(lr){5-7} \cmidrule(lr){8-10} \cmidrule(lr){11-13}
    \multirow{-2}{*}{\textbf{Method}} & \textbf{Acc{$\uparrow$}} & \textbf{Length{$\downarrow$}} & $\textbf{Ratio}_\textbf{T}${$\downarrow$} & \textbf{Acc}{$\uparrow$} & \textbf{Length}{$\downarrow$} & $\textbf{Ratio}_\textbf{T}${$\downarrow$} & \textbf{Acc}{$\uparrow$} & \textbf{Length}{$\downarrow$} & $\textbf{Ratio}_\textbf{T}${$\downarrow$} & \textbf{Acc}{$\uparrow$} & \textbf{Length}{$\downarrow$} & $\textbf{Ratio}_\textbf{T}${$\downarrow$}  \\ \midrule
   Cold-Start (on) & 80.8 & 21265 & 1.00 & 71.7 & 23791 & 1.00  & 56.2 & 19473 & 1.00 & 82.9 & 2662 & 1.00\\
       + GRPO & 82.5\textcolor{deepgreen}{$\scriptstyle\uparrow{\scriptstyle 2.1\%}$} 
& 21045\textcolor{deepgreen}{$\scriptstyle\downarrow{\scriptstyle 1.0\%}$} 
& 1.00\textcolor{uclablue}{$\scriptstyle{\scriptstyle-0.0\%}$} 
& 76.7\textcolor{deepgreen}{$\scriptstyle\uparrow{\scriptstyle 7\%}$} 
& 22695\textcolor{deepgreen}{$\scriptstyle\downarrow{\scriptstyle 4.6\%}$} 
& 1.00\textcolor{uclablue}{$\scriptstyle{\scriptstyle-0.0\%}$}  
& 57.3\textcolor{deepgreen}{$\scriptstyle\uparrow{\scriptstyle 2.0\%}$} 
& 19067\textcolor{deepgreen}{$\scriptstyle\downarrow{\scriptstyle 2.1\%}$} 
& 1.00\textcolor{uclablue}{$\scriptstyle{\scriptstyle -0.0\%}$} 
& \underline{95.1}\textcolor{deepgreen}{$\scriptstyle\uparrow{\scriptstyle 14.7\%}$} 
& 3597\textcolor{deepred}{$\scriptstyle\uparrow{\scriptstyle 35.1\%}$} 
& 1.00\textcolor{uclablue}{$\scriptstyle{\scriptstyle -0.0\%}$} \\

    \midrule
Cold-Start &  85.8\textcolor{deepgreen}{$\scriptstyle\uparrow{\scriptstyle 6.2\%}$} 
& 18138\textcolor{deepgreen}{$\scriptstyle\downarrow{\scriptstyle 14.7\%}$} 
& 1.00\textcolor{uclablue}{$\scriptstyle{\scriptstyle -0.0\%}$} 
& 76.7\textcolor{deepgreen}{$\scriptstyle\uparrow{\scriptstyle 7.0\%}$} 
& 20613\textcolor{deepgreen}{$\scriptstyle\downarrow{\scriptstyle 13.4\%}$} 
& 1.00\textcolor{uclablue}{$\scriptstyle{\scriptstyle-0.0\%}$}  
& 60.8\textcolor{deepgreen}{$\scriptstyle\uparrow{\scriptstyle 8.2\%}$} 
& 18158\textcolor{deepgreen}{$\scriptstyle\downarrow{\scriptstyle 6.8\%}$} 
& 0.91\textcolor{deepgreen}{$\scriptstyle\downarrow{\scriptstyle 9.0\%}$} 
& 88.4\textcolor{deepgreen}{$\scriptstyle\uparrow{\scriptstyle 6.6\%}$} 
& 2272\textcolor{deepgreen}{$\scriptstyle\downarrow{\scriptstyle 14.6\%}$} 
& 0.54\textcolor{deepgreen}{$\scriptstyle\downarrow{\scriptstyle 46.3\%}$} \\

    + GRPO & 86.7\textcolor{deepgreen}{$\scriptstyle\uparrow{\scriptstyle7.2\%}$}  
& 17083\textcolor{deepgreen}{$\scriptstyle\downarrow{\scriptstyle 19.7\%}$} 
& 1.00\textcolor{uclablue}{$\scriptstyle{\scriptstyle-0.0\%}$}  
& 79.17\textcolor{deepgreen}{$\scriptstyle\uparrow{\scriptstyle 10.5\%}$} 
& 19869\textcolor{deepgreen}{$\scriptstyle\downarrow{\scriptstyle 16.5\%}$} 
& 1.00\textcolor{uclablue}{$\scriptstyle{\scriptstyle-0.0\%}$}  
& 62.1\textcolor{deepgreen}{$\scriptstyle\uparrow{\scriptstyle 10.6\%}$} 
& 18046\textcolor{deepgreen}{$\scriptstyle\downarrow{\scriptstyle 7.3\%}$} 
& 0.93\textcolor{deepgreen}{$\scriptstyle\downarrow{\scriptstyle 7.3\%}$} 
& 87.8\textcolor{deepgreen}{$\scriptstyle\uparrow{\scriptstyle 5.9\%}$} 
& 2220\textcolor{deepgreen}{$\scriptstyle\downarrow{\scriptstyle 16.6\%}$} 
& 0.59\textcolor{deepgreen}{$\scriptstyle\downarrow{\scriptstyle 40.8\%}$} \\

    \midrule
    AdaptThink& 83.3\textcolor{deepgreen}{$\scriptstyle\uparrow{\scriptstyle 3.1\%}$} & 16598\textcolor{deepgreen}{$\scriptstyle\downarrow{\scriptstyle 21.9\%}$} & 0.93\textcolor{deepgreen}{$\scriptstyle\downarrow{\scriptstyle 7.0\%}$} & 74.2\textcolor{deepgreen}{$\scriptstyle\uparrow{\scriptstyle 3.5\%}$} & 19993\textcolor{deepgreen}{$\scriptstyle\downarrow{\scriptstyle 16.0\%}$} & 0.84\textcolor{deepgreen}{$\scriptstyle\downarrow{\scriptstyle 16.0\%}$} & 57.1\textcolor{deepgreen}{$\scriptstyle\uparrow{\scriptstyle 1.6\%}$} & 16162\textcolor{deepgreen}{$\scriptstyle\downarrow{\scriptstyle 17.0\%}$} & 0.78\textcolor{deepgreen}{$\scriptstyle\downarrow{\scriptstyle 28.0\%}$} & 85.4\textcolor{deepgreen}{$\scriptstyle\uparrow{\scriptstyle 3.0\%}$} & 915\textcolor{deepgreen}{$\scriptstyle\downarrow{\scriptstyle 65.6\%}$} & 0.16\textcolor{deepgreen}{$\scriptstyle\downarrow{\scriptstyle 84.0\%}$}\\   
    AutoThink& 84.3\textcolor{deepgreen}{$\scriptstyle\uparrow{\scriptstyle 3.5\%}$} & 17061\textcolor{deepgreen}{$\scriptstyle\downarrow{\scriptstyle 19.8\%}$} & 0.95\textcolor{deepgreen}{$\scriptstyle\downarrow{\scriptstyle 5.0\%}$} & 75.0\textcolor{deepgreen}{$\scriptstyle\uparrow{\scriptstyle 4.6\%}$} & 18784\textcolor{deepgreen}{$\scriptstyle\downarrow{\scriptstyle 21.0\%}$} & 0.88\textcolor{deepgreen}{$\scriptstyle\downarrow{\scriptstyle 12.0\%}$} & 57.5\textcolor{deepgreen}{$\scriptstyle\uparrow{\scriptstyle 2.3\%}$} & 15672\textcolor{deepgreen}{$\scriptstyle\downarrow{\scriptstyle 19.5\%}$} & 0.80\textcolor{deepgreen}{$\scriptstyle\downarrow{\scriptstyle 20.0\%}$} & 82.3\textcolor{deepred}{$\scriptstyle\downarrow{\scriptstyle 0.7\%}$} & 1050\textcolor{deepgreen}{$\scriptstyle\downarrow{\scriptstyle 60.6\%}$} & 0.18\textcolor{deepgreen}{$\scriptstyle\downarrow{\scriptstyle 82.0\%}$}\\ 
    HiPO& \underline{87.5}\textcolor{deepgreen}{$\scriptstyle\uparrow{\scriptstyle 8.3\%}$} 
& \underline{15107}\textcolor{deepgreen}{$\scriptstyle\downarrow{\scriptstyle 29.0\%}$} 
& \underline{0.98}\textcolor{deepgreen}{$\scriptstyle\downarrow{\scriptstyle 1.7\%}$} 
& \underline{82.5}\textcolor{deepgreen}{$\scriptstyle\uparrow{\scriptstyle 15.1\%}$} 
& \underline{17655}\textcolor{deepgreen}{$\scriptstyle\downarrow{\scriptstyle 25.8\%}$} 
& \underline{0.95}\textcolor{deepgreen}{$\scriptstyle\downarrow{\scriptstyle 5.0\%}$} 
& \underline{63.0}\textcolor{deepgreen}{$\scriptstyle\uparrow{\scriptstyle 12.2\%}$} 
& \underline{13558}\textcolor{deepgreen}{$\scriptstyle\downarrow{\scriptstyle 30.4\%}$} 
& \underline{0.82}\textcolor{deepgreen}{$\scriptstyle\downarrow{\scriptstyle 18.5\%}$} 
& 90.2\textcolor{deepgreen}{$\scriptstyle\uparrow{\scriptstyle 8.8\%}$} 
& \underline{776}\textcolor{deepgreen}{$\scriptstyle\downarrow{\scriptstyle 70.9\%}$} 
& \underline{0.12}\textcolor{deepgreen}{$\scriptstyle\downarrow{\scriptstyle 88.4\%}$}\\

    \midrule
     & \multicolumn{3}{c|}{\cellcolor{gray!10}\textbf{MATH-500}} & \multicolumn{3}{c|}{\cellcolor{gray!10}\textbf{GPQA-Diamond}} & \multicolumn{3}{c|}{\cellcolor{gray!10}\textbf{MBPP}} & \multicolumn{3}{c}{\cellcolor{gray!10}\textbf{Average}}  \\
      \cmidrule(lr){2-4} \cmidrule(lr){5-7} \cmidrule(lr){8-10} \cmidrule(lr){11-13}
    \multirow{-2}{*}{\textbf{Method}} & \textbf{Acc}{$\uparrow$} & \textbf{Length}{$\downarrow$} & $\textbf{Ratio}_\textbf{T}${$\downarrow$} & \textbf{Acc}{$\uparrow$} & \textbf{Length}{$\downarrow$} & $\textbf{Ratio}_\textbf{T}${$\downarrow$} & \textbf{Acc}{$\uparrow$} & \textbf{Length}{$\downarrow$} & $\textbf{Ratio}_\textbf{T}${$\downarrow$} & \textbf{Acc}{$\uparrow$} & \textbf{Length}{$\downarrow$} & $\textbf{Ratio}_\textbf{T}${$\downarrow$}  \\ \midrule
   Cold-Start (on) & 92.0
& 6237 
& 1.00 
& 61.1 
& 10832
& 1.00
& 72.0  & 4411  & 1.00  
& 73.8 
& 12667
& 1.00 \\

       + GRPO & 93.2\textcolor{deepgreen}{$\scriptstyle\uparrow{\scriptstyle 0.0\%}$} 
& 6256\textcolor{deepred}{$\scriptstyle\uparrow{\scriptstyle 1.3\%}$} 
& 1.00\textcolor{uclablue}{$\scriptstyle{\scriptstyle-0.0\%}$} 
& 57.6\textcolor{deepred}{$\scriptstyle\downarrow{\scriptstyle 5.8\%}$} 
& 10633\textcolor{deepgreen}{$\scriptstyle\downarrow{\scriptstyle 1.8\%}$} 
& 1.00\textcolor{uclablue}{$\scriptstyle{\scriptstyle -0.0\%}$} 
& 71.8\textcolor{deepred}{$\scriptstyle\downarrow{\scriptstyle 0.3\%}$} & 5103\textcolor{deepred}{$\scriptstyle\uparrow{\scriptstyle 15.7\%}$} & 1.00\textcolor{uclablue}{$\scriptstyle-{\scriptstyle 0.0\%}$} 
& 76.3\textcolor{deepgreen}{$\scriptstyle\uparrow{\scriptstyle 3.3\%}$}
& 12628\textcolor{deepgreen}{$\scriptstyle\downarrow{\scriptstyle 0.3\%}$}
& 1.00\textcolor{uclablue}{$\scriptstyle{\scriptstyle -0.0\%}$} \\

    \midrule
Cold-Start & 93.0\textcolor{deepgreen}{$\scriptstyle\uparrow{\scriptstyle 1.1\%}$} 
& 5215\textcolor{deepgreen}{$\scriptstyle\downarrow{\scriptstyle 16.4\%}$} 
& 0.65\textcolor{deepgreen}{$\scriptstyle\downarrow{\scriptstyle 35.0\%}$} 
& \underline{61.6}\textcolor{deepgreen}{$\scriptstyle\uparrow{\scriptstyle 0.8\%}$} 
& 11172\textcolor{deepred}{$\scriptstyle\uparrow{\scriptstyle 3.1\%}$} 
& 0.95\textcolor{deepgreen}{$\scriptstyle\downarrow{\scriptstyle 4.5\%}$} 
& 71.4\textcolor{deepred}{$\scriptstyle\downarrow{\scriptstyle 0.8\%}$} & 3561\textcolor{deepgreen}{$\scriptstyle\downarrow{\scriptstyle 19.3\%}$} & 0.42\textcolor{deepgreen}{$\scriptstyle\downarrow{\scriptstyle 58.0\%}$} 
& 76.8\textcolor{deepgreen}{$\scriptstyle\uparrow{\scriptstyle 4.1\%}$}
& 11304\textcolor{deepgreen}{$\scriptstyle\downarrow{\scriptstyle 10.8\%}$}
& 0.78\textcolor{deepgreen}{$\scriptstyle\downarrow{\scriptstyle 21.8\%}$} \\
    + GRPO & 92.8\textcolor{deepgreen}{$\scriptstyle\uparrow{\scriptstyle 0.9\%}$} 
     & 5204\textcolor{deepgreen}{$\scriptstyle\downarrow{\scriptstyle 16.6\%}$}  
     & 0.68\textcolor{deepgreen}{$\scriptstyle\downarrow{\scriptstyle 31.8\%}$} 
     & 58.6\textcolor{deepred}{$\scriptstyle\downarrow{\scriptstyle 4.1\%}$} 
     & 10581\textcolor{deepgreen}{$\scriptstyle\downarrow{\scriptstyle 2.3\%}$} 
     & 0.98\textcolor{deepgreen}{$\scriptstyle\downarrow{\scriptstyle 2.0\%}$} 
     & 72.0\textcolor{uclablue}{$\scriptstyle{\scriptstyle -0.0\%}$} & 4341\textcolor{deepgreen}{$\scriptstyle\downarrow{\scriptstyle 1.6\%}$} & 0.38\textcolor{deepgreen}{$\scriptstyle\downarrow{\scriptstyle 62.0\%}$} 
     & 77.0\textcolor{deepgreen}{$\scriptstyle\uparrow{\scriptstyle 4.3\%}$}
     & 11049 \textcolor{deepgreen}{$\scriptstyle\downarrow{\scriptstyle 12.8\%}$}
     & 0.79\textcolor{deepgreen}{$\scriptstyle\downarrow{\scriptstyle 20.6\%}$} \\
    \midrule
    AdaptThink& 92.8\textcolor{deepgreen}{$\scriptstyle\uparrow{\scriptstyle 0.9\%}$} & 4213\textcolor{deepgreen}{$\scriptstyle\downarrow{\scriptstyle 32.5\%}$} & 0.55\textcolor{deepgreen}{$\scriptstyle\downarrow{\scriptstyle 45.0\%}$} & 56.1\textcolor{deepred}{$\scriptstyle\downarrow{\scriptstyle 8.2\%}$} & 10242\textcolor{deepgreen}{$\scriptstyle\downarrow{\scriptstyle 5.4\%}$} & 0.91\textcolor{deepgreen}{$\scriptstyle\downarrow{\scriptstyle 9.0\%}$} & 68.0\textcolor{deepred}{$\scriptstyle\downarrow{\scriptstyle 5.6\%}$} & 4165\textcolor{deepgreen}{$\scriptstyle\downarrow{\scriptstyle 5.6\%}$} & 0.33\textcolor{deepgreen}{$\scriptstyle\downarrow{\scriptstyle 67.0\%}$} & 73.8\textcolor{uclablue}{$\scriptstyle-{\scriptstyle 0.0\%}$} & 10327\textcolor{deepgreen}{$\scriptstyle\downarrow{\scriptstyle 18.5\%}$} & 0.64\textcolor{deepgreen}{$\scriptstyle\downarrow{\scriptstyle 36.0\%}$}\\ 
    AutoThink& 92.8\textcolor{deepgreen}{$\scriptstyle\uparrow{\scriptstyle 0.9\%}$} & 4261\textcolor{deepgreen}{$\scriptstyle\downarrow{\scriptstyle 31.7\%}$} & 0.56\textcolor{deepgreen}{$\scriptstyle\downarrow{\scriptstyle 44.0\%}$} & 58.1\textcolor{deepred}{$\scriptstyle\downarrow{\scriptstyle 4.9\%}$} & 9898\textcolor{deepgreen}{$\scriptstyle\downarrow{\scriptstyle 8.6\%}$} & \underline{0.89}\textcolor{deepgreen}{$\scriptstyle\downarrow{\scriptstyle 11.0\%}$} & 70.0\textcolor{deepred}{$\scriptstyle\downarrow{\scriptstyle 2.8\%}$} & 4958\textcolor{deepgreen}{$\scriptstyle\downarrow{\scriptstyle 12.4\%}$} & 0.40\textcolor{deepgreen}{$\scriptstyle\downarrow{\scriptstyle 60.0\%}$} & 74.3\textcolor{deepgreen}{$\scriptstyle\uparrow{\scriptstyle 0.7\%}$} & 10240\textcolor{deepgreen}{$\scriptstyle\downarrow{\scriptstyle 19.2\%}$} & 0.67\textcolor{deepgreen}{$\scriptstyle\downarrow{\scriptstyle 33.0\%}$}\\ 
    HiPO& \underline{93.6} \textcolor{deepgreen}{$\scriptstyle\uparrow{\scriptstyle 1.7\%}$} & \underline{4090} \textcolor{deepgreen}{$\scriptstyle\downarrow{\scriptstyle 34.4\%}$} & \underline{0.54} \textcolor{deepgreen}{$\scriptstyle\downarrow{\scriptstyle 45.8\%}$} & 60.1 \textcolor{deepred}{$\scriptstyle\downarrow{\scriptstyle 1.7\%}$} & \underline{9367} \textcolor{deepgreen}{$\scriptstyle\downarrow{\scriptstyle 13.5\%}$} & 0.92\textcolor{deepgreen}{$\scriptstyle\downarrow{\scriptstyle -8.1\%}$} & \underline{72.2}\textcolor{deepgreen}{$\scriptstyle\uparrow{\scriptstyle 0.3\%}$} & \underline{1338}\textcolor{deepgreen}{$\scriptstyle\downarrow{\scriptstyle 69.7\%}$} & \underline{0.12}\textcolor{deepgreen}{$\scriptstyle\downarrow{\scriptstyle 88.0\%}$} & \underline{78.4}\textcolor{deepgreen}{$\scriptstyle\uparrow{\scriptstyle 6.3\%}$} & \underline{8842}\textcolor{deepgreen}{$\scriptstyle\downarrow{\scriptstyle 30.2\%}$}  & \underline{0.63}\textcolor{deepgreen}{$\scriptstyle\downarrow{\scriptstyle 36.5\%}$} \\ 
    \bottomrule
    \end{tabular}
 }
\caption{Based on Qwen3-8B, performance of different methods on multiple benchmarks. $\textbf{Ratio}_\textbf{T}$ denotes the ratio of ``Think-on'' mode over the corresponding benchmark.}
\label{tab:main}
\end{table}

%% file: Tables/RL_Ablation.tex
\begin{table}[t]
\centering

\resizebox{0.95\linewidth}{!}{
\begin{tabular}{l|ccc|ccc|ccc}
\toprule
& \multicolumn{3}{c|}{\cellcolor{gray!10}\textbf{AIME2024}} & \multicolumn{3}{c|}{\cellcolor{gray!10}\textbf{LiveCodeBench}}  & \multicolumn{3}{c}{\cellcolor{gray!10}\textbf{HumanEval}}\\
\cmidrule(lr){2-10} 
\multirow{-2}{*}{\textbf{Method}} & \textbf{Acc} & \textbf{Length} &$\textbf{Ratio}_\textbf{T}$  &\textbf{Acc} & \textbf{Length} &$\textbf{Ratio}_\textbf{T}$  & \textbf{Acc} & \textbf{Length} &$\textbf{Ratio}_\textbf{T}$ \\ 
\midrule
HiPO & \underline{87.50}&\underline{15107}&\underline{0.98}&\underline{63.00}&\underline{13558}&\underline{0.82}&\underline{90.2}&\underline{776}& \underline{0.12}\\
\midrule
HiPO (w/o global adv)  & 85.83\textcolor{deepred}{$\scriptstyle\downarrow{\scriptstyle 1.9\%}$} & 18064\textcolor{deepred}{$\scriptstyle\uparrow{\scriptstyle 19.6\%}$} & 1.00\textcolor{deepred}{$\scriptstyle\uparrow{\scriptstyle 2.0\%}$} & 56.83\textcolor{deepred}{$\scriptstyle\downarrow{\scriptstyle 9.8\%}$} & 14561\textcolor{deepred}{$\scriptstyle\uparrow{\scriptstyle 7.4\%}$} & 0.86\textcolor{deepred}{$\scriptstyle\uparrow{\scriptstyle 4.9\%}$} & 89.63\textcolor{deepred}{$\scriptstyle\downarrow{\scriptstyle 4.9\%}$} & 1660\textcolor{deepred}{$\scriptstyle\uparrow{\scriptstyle 114.9\%}$} & 0.27\textcolor{deepred}{$\scriptstyle\uparrow{\scriptstyle 125.0\%}$} \\
HiPO (w/o local norm) & 85.00\textcolor{deepred}{$\scriptstyle\downarrow{\scriptstyle 2.9\%}$} & 18268\textcolor{deepred}{$\scriptstyle\uparrow{\scriptstyle 20.9\%}$} & 1.00\textcolor{deepred}{$\scriptstyle\uparrow{\scriptstyle 2.0\%}$} & 58.37\textcolor{deepred}{$\scriptstyle\downarrow{\scriptstyle 7.3\%}$} & 16029\textcolor{deepred}{$\scriptstyle\uparrow{\scriptstyle 18.2\%}$} & 0.88\textcolor{deepred}{$\scriptstyle\uparrow{\scriptstyle 7.3\%}$} & 89.63\textcolor{deepred}{$\scriptstyle\downarrow{\scriptstyle 0.6\%}$} & 2052\textcolor{deepred}{$\scriptstyle\uparrow{\scriptstyle 164.4\%}$} & 0.32\textcolor{deepred}{$\scriptstyle\uparrow{\scriptstyle 166.7\%}$}\\
\bottomrule
\end{tabular}
 }
\caption{Performance of different design strategies on advantage functions.}
\label{a_variants}
\end{table}

%% file: content/5_conclusion.tex
\section{Conclusion}
In this work, we introduced HiPO, a hybrid framework for adaptive reasoning in LLMs. By combining a hybrid data pipeline with a hybrid reinforcement learning reward system, HiPO enables models to dynamically balance Think-on and Think-off reasoning, mitigating the issue of overthinking while preserving accuracy. Experiments demonstrate that HiPO achieves competitive or superior accuracy with significantly improved token efficiency and reduced reasoning redundancy.

%% file: content/6_appendix.tex
\clearpage
\section{Appendix}

\subsection{Use of LLMs}
LLMs were used solely to assist in editing, formatting, and improving the clarity of the manuscript. 
All ideas, experiments, and analyses were conceived and executed by the authors. 
No LLM outputs were used as experimental data or results in this work.

\subsection{The decline in Qwen3's performance on the test set.}
\label{qwen3_decline}
\begin{figure}[H]
    \centering
    \includegraphics[width=1.0\linewidth]{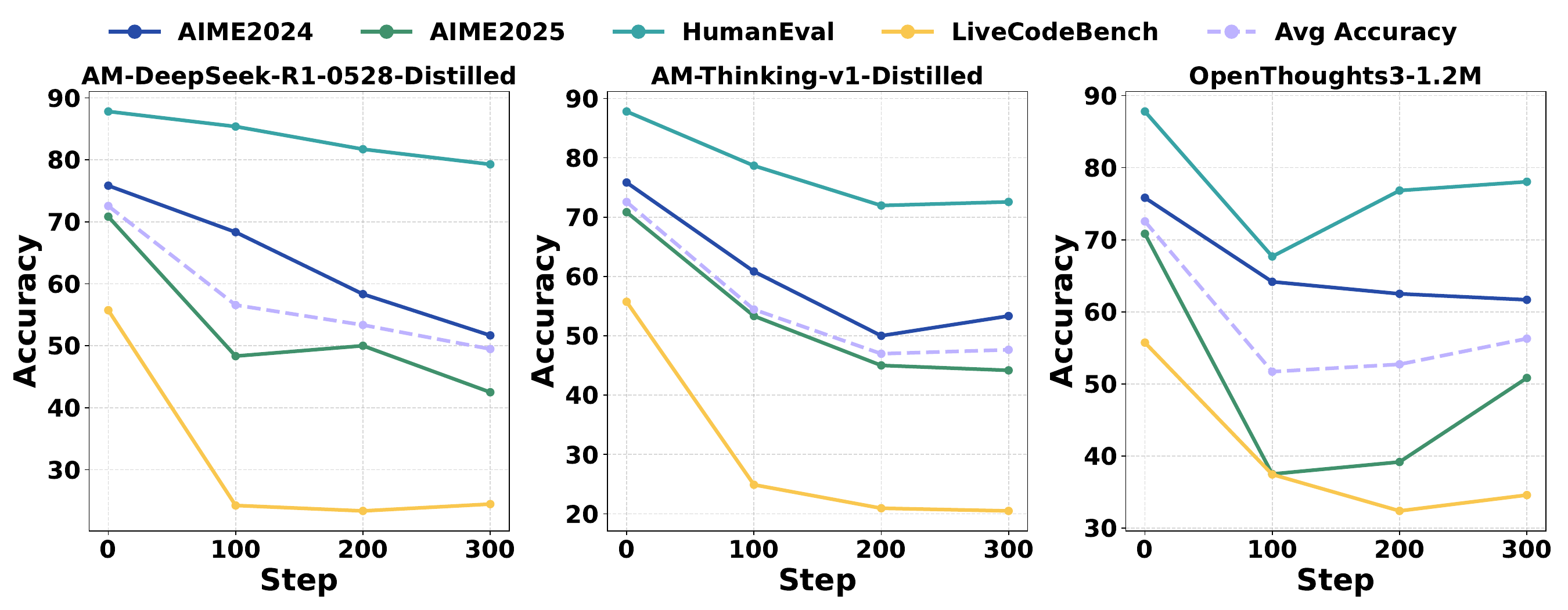}
    \caption{The decline in Qwen3's performance on the AIME2024, AIME2025, HumanEval, LiveCodeBench.}
    \label{fig:placeholder}
\end{figure}

This section demonstrates the decline in Qwen3's performance on AIME2024, AIME2025, HumanEval, and LiverCodeBench. 
We trained Qwen3 using AM-DeepSeek-R1-0528-Distilled, AM-Thinking-v1-Distilled, and OpenThoughts3-1.2M. 
The Figure \ref{fig:placeholder}, when the number of training steps reaches 150, Qwen3's accuracy on all benchmarks declines.
Note that the batch size is set as 512 and other parameters are same as the implementation details in the main paper.


\subsection{Data Source}
\label{app:data_source}
Our dataset is derived from several open-source reasoning corpora covering both code and mathematics. 
As shown in Table~\ref{tab:data_composition}, queries come from AM-Thinking-v1-Distilled~\footnote{https://huggingface.co/datasets/a-m-team/AM-Thinking-v1-Distilled}, II-Thought-RL~\footnote{https://huggingface.co/datasets/Intelligent-Internet/II-Thought-RL-v0}, AceReason-Math~\footnote{https://huggingface.co/datasets/nvidia/AceReason-Math}, and Skywork-OR1-RL-Data~\footnote{https://huggingface.co/datasets/Skywork/Skywork-OR1-RL-Data}. 
This composition ensures diversity across domains and provides a reliable basis for model training and evaluation.

\begin{table}[tp]
    \centering
    \renewcommand{\arraystretch}{1.1}
    \resizebox{0.8\textwidth}{!}{
        \begin{tabular}{c|c|c}
        \toprule
        \textbf{Category} & \textbf{Data Source} & \textbf{\# Query} \\
        \midrule
        \multirow{2}{*}{Code} 
            & AM-Thinking-v1-Distilled~\citep{tian2025not} & 85k \\
            & II-Thought-RL~\citep{2025iithought} & 20k \\
        \midrule
        \multirow{4}{*}{Math} 
            & AceReason-Math~\citep{chen2025acereason} & 49k \\
            & AM-Thinking-v1-Distilled~\citep{tian2025not} & 32k \\
            & II-Thought-RL~\citep{2025iithought} & 30k \\
            & Skywork-OR1-RL-Data~\citep{skywork-or1-2025} & 24k \\
        \bottomrule
        \end{tabular}
    }
        \caption{Description of data sources.}

    \label{tab:data_composition}
\end{table}

\subsection{Prompt Templates}
\label{app:prompt_templates_for_tasks}
In this section, we provide the prompt templates for the response generation and judge analysis generation.

\begin{tcolorbox}[title = {Response Generation}]
Please read the following question carefully and provide a clear answer.

---

{\color{blue} Query}

---

\end{tcolorbox}

\begin{tcolorbox}[title = {Judge Analysis Generation}]
You are tasked with analyzing the characteristics of a question to determine why it **requires** complex reasoning.

Your should **not** attempting to answer or infer its solution. 

You should analyse user's question to determine the **core task intention**—that is, what the user wants the model to do. (e.g., write and validate code based on a problem description, etc.). 

Then briefly outline the basic approach to accomplishing this task (e.g., write SQL code to retrieve imformation, etc.). 

Based on the required approach, assess the **reasoning complexity**, and indicate whether it involves multiple steps or deep analysis. Do not solve the question or provide an answer. Focus solely on interpreting the task type, approach, and cognitive demand.

Be concise: your analysis must be no more than two lines and under 500 characters. Use clear, natural, and varied language.
End your explanation with a statement indicating that complex reasoning is required (Think-on), but express this conclusion with a natural and diverse phrase, not repeating any single pattern. The meaning must be clear, but the expression can vary.

Please analyze the following question as required above:

---

{\color{blue} Model Response}

---
\end{tcolorbox}

%% file: main.bbl
\begin{thebibliography}{44}
\providecommand{\natexlab}[1]{#1}
\providecommand{\url}[1]{\texttt{#1}}
\expandafter\ifx\csname urlstyle\endcsname\relax
  \providecommand{\doi}[1]{doi: #1}\else
  \providecommand{\doi}{doi: \begingroup \urlstyle{rm}\Url}\fi

\bibitem[Yao et~al.(2023)Yao, Yu, Zhao, Shafran, Griffiths, Cao, and Narasimhan]{yao2023tree}
Shunyu Yao, Dian Yu, Jeffrey Zhao, Izhak Shafran, Thomas~L. Griffiths, Yuan Cao, and Karthik Narasimhan.
\newblock Tree of thoughts: Deliberate problem solving with large language models.
\newblock \emph{arXiv preprint arXiv: 2305.10601}, 2023.

\bibitem[Wei et~al.(2023)Wei, Wang, Schuurmans, Bosma, Ichter, Xia, Chi, Le, and Zhou]{wei2023chainofthoughtpromptingelicitsreasoning}
Jason Wei, Xuezhi Wang, Dale Schuurmans, Maarten Bosma, Brian Ichter, Fei Xia, Ed~Chi, Quoc Le, and Denny Zhou.
\newblock Chain-of-thought prompting elicits reasoning in large language models, 2023.
\newblock URL \url{https://arxiv.org/abs/2201.11903}.

\bibitem[Kumar et~al.(2025)Kumar, Roh, Naseh, Karpinska, Iyyer, Houmansadr, and Bagdasarian]{kumar2025overthinkslowdownattacksreasoning}
Abhinav Kumar, Jaechul Roh, Ali Naseh, Marzena Karpinska, Mohit Iyyer, Amir Houmansadr, and Eugene Bagdasarian.
\newblock Overthink: Slowdown attacks on reasoning llms, 2025.
\newblock URL \url{https://arxiv.org/abs/2502.02542}.

\bibitem[Sui et~al.(2025)Sui, Chuang, Wang, Zhang, Zhang, Yuan, Liu, Wen, Zhong, Zou, Chen, and Hu]{sui2025stopoverthinkingsurveyefficient}
Yang Sui, Yu-Neng Chuang, Guanchu Wang, Jiamu Zhang, Tianyi Zhang, Jiayi Yuan, Hongyi Liu, Andrew Wen, Shaochen Zhong, Na~Zou, Hanjie Chen, and Xia Hu.
\newblock Stop overthinking: A survey on efficient reasoning for large language models, 2025.
\newblock URL \url{https://arxiv.org/abs/2503.16419}.

\bibitem[Nayab et~al.(2025)Nayab, Rossolini, Simoni, Saracino, Buttazzo, Manes, and Giacomelli]{nayab2025concisethoughtsimpactoutput}
Sania Nayab, Giulio Rossolini, Marco Simoni, Andrea Saracino, Giorgio Buttazzo, Nicolamaria Manes, and Fabrizio Giacomelli.
\newblock Concise thoughts: Impact of output length on llm reasoning and cost, 2025.
\newblock URL \url{https://arxiv.org/abs/2407.19825}.

\bibitem[Aggarwal and Welleck(2025)]{aggarwal2025l1controllinglongreasoning}
Pranjal Aggarwal and Sean Welleck.
\newblock L1: Controlling how long a reasoning model thinks with reinforcement learning, 2025.
\newblock URL \url{https://arxiv.org/abs/2503.04697}.

\bibitem[Arora and Zanette(2025)]{arora2025traininglanguagemodelsreason}
Daman Arora and Andrea Zanette.
\newblock Training language models to reason efficiently, 2025.
\newblock URL \url{https://arxiv.org/abs/2502.04463}.

\bibitem[Hou et~al.(2025)Hou, Zhang, Ji, Liu, Qian, Andreas, and Chang]{hou2025thinkprunepruninglongchainofthought}
Bairu Hou, Yang Zhang, Jiabao Ji, Yujian Liu, Kaizhi Qian, Jacob Andreas, and Shiyu Chang.
\newblock Thinkprune: Pruning long chain-of-thought of llms via reinforcement learning, 2025.
\newblock URL \url{https://arxiv.org/abs/2504.01296}.

\bibitem[Luo et~al.(2025)Luo, Shen, He, Wang, Liu, Li, Tan, Cao, and Tao]{luo2025o1prunerlengthharmonizingfinetuningo1like}
Haotian Luo, Li~Shen, Haiying He, Yibo Wang, Shiwei Liu, Wei Li, Naiqiang Tan, Xiaochun Cao, and Dacheng Tao.
\newblock O1-pruner: Length-harmonizing fine-tuning for o1-like reasoning pruning, 2025.
\newblock URL \url{https://arxiv.org/abs/2501.12570}.

\bibitem[Shen et~al.(2025)Shen, Zhang, Huang, Shi, Zhang, Yan, Wang, Wang, Liu, and Lian]{shen2025dastdifficultyadaptiveslowthinkinglarge}
Yi~Shen, Jian Zhang, Jieyun Huang, Shuming Shi, Wenjing Zhang, Jiangze Yan, Ning Wang, Kai Wang, Zhaoxiang Liu, and Shiguo Lian.
\newblock Dast: Difficulty-adaptive slow-thinking for large reasoning models, 2025.
\newblock URL \url{https://arxiv.org/abs/2503.04472}.

\bibitem[Team et~al.(2025)Team, Du, Gao, Xing, Jiang, Chen, Li, Xiao, Du, Liao, Tang, Wang, Zhang, Yuan, Lu, Tang, Sung, Wei, Lai, Guo, Zhu, Ding, Hu, Yang, Zhang, Yao, Zhao, Lu, Li, Yu, Gao, Zheng, Yuan, Chen, Guo, Su, Wang, Zhao, Zhang, Liu, Yan, Wu, Shi, Ye, Yu, Dong, Zhang, Ma, Pan, Gong, Liu, Ma, Wei, Cao, Huang, Jiang, Gao, Xiong, He, Huang, Xu, Wu, He, Wei, Jia, Wu, Xu, Zu, Zhou, Pan, Charles, Li, Hu, Liu, Chen, Wang, Liu, Qin, Liu, Yang, Bao, Du, Wu, Wang, Zhou, Wang, Li, Zhu, Zhang, Wang, Yang, Huang, Huang, Xu, Yang, and Lin]{kimiteam2025kimik15scalingreinforcement}
Kimi Team, Angang Du, Bofei Gao, Bowei Xing, Changjiu Jiang, Cheng Chen, Cheng Li, Chenjun Xiao, Chenzhuang Du, Chonghua Liao, Chuning Tang, Congcong Wang, Dehao Zhang, Enming Yuan, Enzhe Lu, Fengxiang Tang, Flood Sung, Guangda Wei, Guokun Lai, Haiqing Guo, Han Zhu, Hao Ding, Hao Hu, Hao Yang, Hao Zhang, Haotian Yao, Haotian Zhao, Haoyu Lu, Haoze Li, Haozhen Yu, Hongcheng Gao, Huabin Zheng, Huan Yuan, Jia Chen, Jianhang Guo, Jianlin Su, Jianzhou Wang, Jie Zhao, Jin Zhang, Jingyuan Liu, Junjie Yan, Junyan Wu, Lidong Shi, Ling Ye, Longhui Yu, Mengnan Dong, Neo Zhang, Ningchen Ma, Qiwei Pan, Qucheng Gong, Shaowei Liu, Shengling Ma, Shupeng Wei, Sihan Cao, Siying Huang, Tao Jiang, Weihao Gao, Weimin Xiong, Weiran He, Weixiao Huang, Weixin Xu, Wenhao Wu, Wenyang He, Xianghui Wei, Xianqing Jia, Xingzhe Wu, Xinran Xu, Xinxing Zu, Xinyu Zhou, Xuehai Pan, Y.~Charles, Yang Li, Yangyang Hu, Yangyang Liu, Yanru Chen, Yejie Wang, Yibo Liu, Yidao Qin, Yifeng Liu, Ying Yang, Yiping Bao, Yulun Du, Yuxin Wu, Yuzhi Wang, Zaida
  Zhou, Zhaoji Wang, Zhaowei Li, Zhen Zhu, Zheng Zhang, Zhexu Wang, Zhilin Yang, Zhiqi Huang, Zihao Huang, Ziyao Xu, Zonghan Yang, and Zongyu Lin.
\newblock Kimi k1.5: Scaling reinforcement learning with llms, 2025.
\newblock URL \url{https://arxiv.org/abs/2501.12599}.

\bibitem[Lou et~al.(2025)Lou, Sun, Liang, Qu, Shen, Wang, Li, Yang, and Wu]{lou2025adacotparetooptimaladaptivechainofthought}
Chenwei Lou, Zewei Sun, Xinnian Liang, Meng Qu, Wei Shen, Wenqi Wang, Yuntao Li, Qingping Yang, and Shuangzhi Wu.
\newblock Adacot: Pareto-optimal adaptive chain-of-thought triggering via reinforcement learning, 2025.
\newblock URL \url{https://arxiv.org/abs/2505.11896}.

\bibitem[Munkhdalai et~al.(2024)Munkhdalai, Faruqui, and Gopal]{munkhdalai2024leave}
Tsendsuren Munkhdalai, Manaal Faruqui, and Siddharth Gopal.
\newblock Leave no context behind: Efficient infinite context transformers with infini-attention.
\newblock \emph{arXiv preprint arXiv:2404.07143}, 2024.

\bibitem[Ma et~al.(2025)Ma, Wan, Yu, Fang, and Wang]{ma2025cotvalvelengthcompressiblechainofthoughttuning}
Xinyin Ma, Guangnian Wan, Runpeng Yu, Gongfan Fang, and Xinchao Wang.
\newblock Cot-valve: Length-compressible chain-of-thought tuning, 2025.
\newblock URL \url{https://arxiv.org/abs/2502.09601}.

\bibitem[Chen et~al.(2025{\natexlab{a}})Chen, Xu, Liang, He, Pang, Yu, Song, Liu, Zhou, Zhang, Wang, Tu, Mi, and Yu]{chen2025think23overthinkingo1like}
Xingyu Chen, Jiahao Xu, Tian Liang, Zhiwei He, Jianhui Pang, Dian Yu, Linfeng Song, Qiuzhi Liu, Mengfei Zhou, Zhuosheng Zhang, Rui Wang, Zhaopeng Tu, Haitao Mi, and Dong Yu.
\newblock Do not think that much for 2+3=? on the overthinking of o1-like llms, 2025{\natexlab{a}}.
\newblock URL \url{https://arxiv.org/abs/2412.21187}.

\bibitem[Kang et~al.(2025)Kang, Sun, Chen, and Zou]{10.1609/aaai.v39i23.34608}
Yu~Kang, Xianghui Sun, Liangyu Chen, and Wei Zou.
\newblock C3ot: generating shorter chain-of-thought without compromising effectiveness.
\newblock In \emph{Proceedings of the Thirty-Ninth AAAI Conference on Artificial Intelligence and Thirty-Seventh Conference on Innovative Applications of Artificial Intelligence and Fifteenth Symposium on Educational Advances in Artificial Intelligence}, AAAI'25/IAAI'25/EAAI'25. AAAI Press, 2025.
\newblock ISBN 978-1-57735-897-8.
\newblock \doi{10.1609/aaai.v39i23.34608}.
\newblock URL \url{https://doi.org/10.1609/aaai.v39i23.34608}.

\bibitem[Xu et~al.(2025)Xu, Xie, Zhao, and He]{xu2025chaindraftthinkingfaster}
Silei Xu, Wenhao Xie, Lingxiao Zhao, and Pengcheng He.
\newblock Chain of draft: Thinking faster by writing less, 2025.
\newblock URL \url{https://arxiv.org/abs/2502.18600}.

\bibitem[Renze and Guven(2024)]{Renze_2024}
Matthew Renze and Erhan Guven.
\newblock The benefits of a concise chain of thought on problem-solving in large language models.
\newblock In \emph{2024 2nd International Conference on Foundation and Large Language Models (FLLM)}, page 476–483. IEEE, November 2024.
\newblock \doi{10.1109/fllm63129.2024.10852493}.
\newblock URL \url{http://dx.doi.org/10.1109/FLLM63129.2024.10852493}.

\bibitem[Chen et~al.(2024)Chen, Qin, Wang, Zhou, and Che]{chen2024unlockingcapabilitiesthoughtreasoning}
Qiguang Chen, Libo Qin, Jiaqi Wang, Jinxuan Zhou, and Wanxiang Che.
\newblock Unlocking the capabilities of thought: A reasoning boundary framework to quantify and optimize chain-of-thought, 2024.
\newblock URL \url{https://arxiv.org/abs/2410.05695}.

\bibitem[Munkhbat et~al.(2025)Munkhbat, Ho, Kim, Yang, Kim, and Yun]{munkhbat2025selftrainingelicitsconcisereasoning}
Tergel Munkhbat, Namgyu Ho, Seo~Hyun Kim, Yongjin Yang, Yujin Kim, and Se-Young Yun.
\newblock Self-training elicits concise reasoning in large language models, 2025.
\newblock URL \url{https://arxiv.org/abs/2502.20122}.

\bibitem[Liu et~al.(2024{\natexlab{a}})Liu, Feng, Xue, Wang, Wu, Lu, Zhao, Deng, Zhang, Ruan, et~al.]{liu2024deepseek}
Aixin Liu, Bei Feng, Bing Xue, Bingxuan Wang, Bochao Wu, Chengda Lu, Chenggang Zhao, Chengqi Deng, Chenyu Zhang, Chong Ruan, et~al.
\newblock Deepseek-v3 technical report.
\newblock \emph{arXiv preprint arXiv:2412.19437}, 2024{\natexlab{a}}.

\bibitem[Shao et~al.(2024)Shao, Wang, Zhu, Xu, Song, Bi, Zhang, Zhang, Li, Wu, and Guo]{shao2024deepseekmath}
Zhihong Shao, Peiyi Wang, Qihao Zhu, Runxin Xu, Junxiao Song, Xiao Bi, Haowei Zhang, Mingchuan Zhang, Y.~K. Li, Y.~Wu, and Daya Guo.
\newblock Deepseekmath: Pushing the limits of mathematical reasoning in open language models.
\newblock \emph{arXiv preprint arXiv: 2402.03300}, 2024.
\newblock URL \url{https://arxiv.org/abs/2402.03300v3}.

\bibitem[Zheng et~al.(2025)Zheng, Liu, Li, Chen, Yu, Gao, Dang, Liu, Men, Yang, Zhou, and Lin]{zheng2025groupsequencepolicyoptimization}
Chujie Zheng, Shixuan Liu, Mingze Li, Xiong-Hui Chen, Bowen Yu, Chang Gao, Kai Dang, Yuqiong Liu, Rui Men, An~Yang, Jingren Zhou, and Junyang Lin.
\newblock Group sequence policy optimization, 2025.
\newblock URL \url{https://arxiv.org/abs/2507.18071}.

\bibitem[Yue et~al.(2025)Yue, Yuan, Yu, Zuo, Zhu, Xu, Chen, Wang, Fan, Du, Wei, Yu, Liu, Liu, Liu, Lin, Lin, Ma, Zhang, Zhang, Zhang, Zhu, Zhang, Liu, Wang, Wu, and Yan]{yue2025vapoefficientreliablereinforcement}
Yu~Yue, Yufeng Yuan, Qiying Yu, Xiaochen Zuo, Ruofei Zhu, Wenyuan Xu, Jiaze Chen, Chengyi Wang, TianTian Fan, Zhengyin Du, Xiangpeng Wei, Xiangyu Yu, Gaohong Liu, Juncai Liu, Lingjun Liu, Haibin Lin, Zhiqi Lin, Bole Ma, Chi Zhang, Mofan Zhang, Wang Zhang, Hang Zhu, Ru~Zhang, Xin Liu, Mingxuan Wang, Yonghui Wu, and Lin Yan.
\newblock Vapo: Efficient and reliable reinforcement learning for advanced reasoning tasks, 2025.
\newblock URL \url{https://arxiv.org/abs/2504.05118}.

\bibitem[Schulman et~al.(2017)Schulman, Wolski, Dhariwal, Radford, and Klimov]{schulman2017proximalpolicyoptimizationalgorithms}
John Schulman, Filip Wolski, Prafulla Dhariwal, Alec Radford, and Oleg Klimov.
\newblock Proximal policy optimization algorithms, 2017.
\newblock URL \url{https://arxiv.org/abs/1707.06347}.

\bibitem[Rafailov et~al.(2024)Rafailov, Sharma, Mitchell, Ermon, Manning, and Finn]{rafailov2024directpreferenceoptimizationlanguage}
Rafael Rafailov, Archit Sharma, Eric Mitchell, Stefano Ermon, Christopher~D. Manning, and Chelsea Finn.
\newblock Direct preference optimization: Your language model is secretly a reward model, 2024.
\newblock URL \url{https://arxiv.org/abs/2305.18290}.

\bibitem[DeepSeek-AI et~al.(2025)DeepSeek-AI, Guo, Yang, Zhang, Song, Zhang, Xu, Zhu, Ma, Wang, Bi, Zhang, Yu, Wu, Wu, Gou, Shao, Li, Gao, Liu, Xue, Wang, Wu, Feng, Lu, Zhao, Deng, Zhang, Ruan, Dai, Chen, Ji, Li, Lin, Dai, Luo, Hao, Chen, Li, Zhang, Bao, Xu, Wang, Ding, Xin, Gao, Qu, Li, Guo, Li, Wang, Chen, Yuan, Qiu, Li, Cai, Ni, Liang, Chen, Dong, Hu, Gao, Guan, Huang, Yu, Wang, Zhang, Zhao, Wang, Zhang, Xu, Xia, Zhang, Zhang, Tang, Li, Wang, Li, Tian, Huang, Zhang, Wang, Chen, Du, Ge, Zhang, Pan, Wang, Chen, Jin, Chen, Lu, Zhou, Chen, Ye, Wang, Yu, Zhou, Pan, Li, Zhou, Wu, Ye, Yun, Pei, Sun, Wang, Zeng, Zhao, Liu, Liang, Gao, Yu, Zhang, Xiao, An, Liu, Wang, Chen, Nie, Cheng, Liu, Xie, Liu, Yang, Li, Su, Lin, Li, Jin, Shen, Chen, Sun, Wang, Song, Zhou, Wang, Shan, Li, Wang, Wei, Zhang, Xu, Li, Zhao, Sun, Wang, Yu, Zhang, Shi, Xiong, He, Piao, Wang, Tan, Ma, Liu, Guo, Ou, Wang, Gong, Zou, He, Xiong, Luo, You, Liu, Zhou, Zhu, Xu, Huang, Li, Zheng, Zhu, Ma, Tang, Zha, Yan, Ren, Ren, Sha, Fu, Xu, Xie, Zhang,
  Hao, Ma, Yan, Wu, Gu, Zhu, Liu, Li, Xie, Song, Pan, Huang, Xu, Zhang, and Zhang]{deepseekai2025deepseekr1incentivizingreasoningcapability}
DeepSeek-AI, Daya Guo, Dejian Yang, Haowei Zhang, Junxiao Song, Ruoyu Zhang, Runxin Xu, Qihao Zhu, Shirong Ma, Peiyi Wang, Xiao Bi, Xiaokang Zhang, Xingkai Yu, Yu~Wu, Z.~F. Wu, Zhibin Gou, Zhihong Shao, Zhuoshu Li, Ziyi Gao, Aixin Liu, Bing Xue, Bingxuan Wang, Bochao Wu, Bei Feng, Chengda Lu, Chenggang Zhao, Chengqi Deng, Chenyu Zhang, Chong Ruan, Damai Dai, Deli Chen, Dongjie Ji, Erhang Li, Fangyun Lin, Fucong Dai, Fuli Luo, Guangbo Hao, Guanting Chen, Guowei Li, H.~Zhang, Han Bao, Hanwei Xu, Haocheng Wang, Honghui Ding, Huajian Xin, Huazuo Gao, Hui Qu, Hui Li, Jianzhong Guo, Jiashi Li, Jiawei Wang, Jingchang Chen, Jingyang Yuan, Junjie Qiu, Junlong Li, J.~L. Cai, Jiaqi Ni, Jian Liang, Jin Chen, Kai Dong, Kai Hu, Kaige Gao, Kang Guan, Kexin Huang, Kuai Yu, Lean Wang, Lecong Zhang, Liang Zhao, Litong Wang, Liyue Zhang, Lei Xu, Leyi Xia, Mingchuan Zhang, Minghua Zhang, Minghui Tang, Meng Li, Miaojun Wang, Mingming Li, Ning Tian, Panpan Huang, Peng Zhang, Qiancheng Wang, Qinyu Chen, Qiushi Du, Ruiqi Ge, Ruisong
  Zhang, Ruizhe Pan, Runji Wang, R.~J. Chen, R.~L. Jin, Ruyi Chen, Shanghao Lu, Shangyan Zhou, Shanhuang Chen, Shengfeng Ye, Shiyu Wang, Shuiping Yu, Shunfeng Zhou, Shuting Pan, S.~S. Li, Shuang Zhou, Shaoqing Wu, Shengfeng Ye, Tao Yun, Tian Pei, Tianyu Sun, T.~Wang, Wangding Zeng, Wanjia Zhao, Wen Liu, Wenfeng Liang, Wenjun Gao, Wenqin Yu, Wentao Zhang, W.~L. Xiao, Wei An, Xiaodong Liu, Xiaohan Wang, Xiaokang Chen, Xiaotao Nie, Xin Cheng, Xin Liu, Xin Xie, Xingchao Liu, Xinyu Yang, Xinyuan Li, Xuecheng Su, Xuheng Lin, X.~Q. Li, Xiangyue Jin, Xiaojin Shen, Xiaosha Chen, Xiaowen Sun, Xiaoxiang Wang, Xinnan Song, Xinyi Zhou, Xianzu Wang, Xinxia Shan, Y.~K. Li, Y.~Q. Wang, Y.~X. Wei, Yang Zhang, Yanhong Xu, Yao Li, Yao Zhao, Yaofeng Sun, Yaohui Wang, Yi~Yu, Yichao Zhang, Yifan Shi, Yiliang Xiong, Ying He, Yishi Piao, Yisong Wang, Yixuan Tan, Yiyang Ma, Yiyuan Liu, Yongqiang Guo, Yuan Ou, Yuduan Wang, Yue Gong, Yuheng Zou, Yujia He, Yunfan Xiong, Yuxiang Luo, Yuxiang You, Yuxuan Liu, Yuyang Zhou, Y.~X. Zhu,
  Yanhong Xu, Yanping Huang, Yaohui Li, Yi~Zheng, Yuchen Zhu, Yunxian Ma, Ying Tang, Yukun Zha, Yuting Yan, Z.~Z. Ren, Zehui Ren, Zhangli Sha, Zhe Fu, Zhean Xu, Zhenda Xie, Zhengyan Zhang, Zhewen Hao, Zhicheng Ma, Zhigang Yan, Zhiyu Wu, Zihui Gu, Zijia Zhu, Zijun Liu, Zilin Li, Ziwei Xie, Ziyang Song, Zizheng Pan, Zhen Huang, Zhipeng Xu, Zhongyu Zhang, and Zhen Zhang.
\newblock Deepseek-r1: Incentivizing reasoning capability in llms via reinforcement learning, 2025.
\newblock URL \url{https://arxiv.org/abs/2501.12948}.

\bibitem[Zhan et~al.(2025)Zhan, Deng, Tang, Xiang, Wu, Li, Zhu, Xu, Huang, Feng, Wang, Yan, Chen, Liu, Peng, Gao, Huang, Zhang, Wang, Lin, Li, Wang, Zhan, Wu, Zhang, Yang, Chen, Zhang, Chen, and Yu]{zhan2025katv1kwaiautothinktechnicalreport}
Zizheng Zhan, Ken Deng, Huaixi Tang, Wen Xiang, Kun Wu, Weihao Li, Wenqiang Zhu, Jingxuan Xu, Lecheng Huang, Zongxian Feng, Shaojie Wang, Shangpeng Yan, Xuxing Chen, Jiaheng Liu, Zhongyuan Peng, Zuchen Gao, Haoyang Huang, Xiaojiang Zhang, Jinghui Wang, Zheng Lin, Mengtong Li, Huiming Wang, Ziqi Zhan, Yanan Wu, Yuanxing Zhang, Jian Yang, Guang Chen, Haotian Zhang, Bin Chen, and Bing Yu.
\newblock Kat-v1: Kwai-autothink technical report, 2025.
\newblock URL \url{https://arxiv.org/abs/2507.08297}.

\bibitem[Aytes et~al.(2025)Aytes, Baek, and Hwang]{aytes2025sketchofthoughtefficientllmreasoning}
Simon~A. Aytes, Jinheon Baek, and Sung~Ju Hwang.
\newblock Sketch-of-thought: Efficient llm reasoning with adaptive cognitive-inspired sketching, 2025.
\newblock URL \url{https://arxiv.org/abs/2503.05179}.

\bibitem[Xia et~al.(2025)Xia, Leong, Wang, Li, and Li]{xia2025tokenskipcontrollablechainofthoughtcompression}
Heming Xia, Chak~Tou Leong, Wenjie Wang, Yongqi Li, and Wenjie Li.
\newblock Tokenskip: Controllable chain-of-thought compression in llms, 2025.
\newblock URL \url{https://arxiv.org/abs/2502.12067}.

\bibitem[Liu et~al.(2024{\natexlab{b}})Liu, Guo, Hu, Jiayang, Zhang, Qiu, and Zhang]{liu2024languagemodelslearnskip}
Tengxiao Liu, Qipeng Guo, Xiangkun Hu, Cheng Jiayang, Yue Zhang, Xipeng Qiu, and Zheng Zhang.
\newblock Can language models learn to skip steps?, 2024{\natexlab{b}}.
\newblock URL \url{https://arxiv.org/abs/2411.01855}.

\bibitem[Sun et~al.(2024)Sun, Haider, Zhang, Yang, Qiu, Yin, Wang, Bartlett, and Zanette]{sun2024fastbestofndecodingspeculative}
Hanshi Sun, Momin Haider, Ruiqi Zhang, Huitao Yang, Jiahao Qiu, Ming Yin, Mengdi Wang, Peter Bartlett, and Andrea Zanette.
\newblock Fast best-of-n decoding via speculative rejection, 2024.
\newblock URL \url{https://arxiv.org/abs/2410.20290}.

\bibitem[Yang et~al.(2025)Yang, Si, Duan, Zhu, Zhu, Li, Lin, Cao, and Wang]{yang2025dynamicearlyexitreasoning}
Chenxu Yang, Qingyi Si, Yongjie Duan, Zheliang Zhu, Chenyu Zhu, Qiaowei Li, Zheng Lin, Li~Cao, and Weiping Wang.
\newblock Dynamic early exit in reasoning models, 2025.
\newblock URL \url{https://arxiv.org/abs/2504.15895}.

\bibitem[Tian et~al.(2025)Tian, Ji, Wang, Chen, Zhao, Peng, Zhao, and Li]{tian2025not}
Xiaoyu Tian, Yunjie Ji, Haotian Wang, Shuaiting Chen, Sitong Zhao, Yiping Peng, Han Zhao, and Xiangang Li.
\newblock Not all correct answers are equal: Why your distillation source matters.
\newblock \emph{arXiv preprint arXiv:2505.14464}, 2025.

\bibitem[Internet(2025)]{2025iithought}
Intelligent Internet.
\newblock Ii-thought : A large-scale, high-quality reasoning dataset, 2025.

\bibitem[Chen et~al.(2025{\natexlab{b}})Chen, Yang, Liu, Lee, Xu, Shoeybi, Catanzaro, and Ping]{chen2025acereason}
Yang Chen, Zhuolin Yang, Zihan Liu, Chankyu Lee, Peng Xu, Mohammad Shoeybi, Bryan Catanzaro, and Wei Ping.
\newblock Acereason-nemotron: Advancing math and code reasoning through reinforcement learning.
\newblock \emph{arXiv preprint arXiv:2505.16400}, 2025{\natexlab{b}}.

\bibitem[He et~al.(2025)He, Liu, Liu, Yan, Wang, Cheng, Zhang, Zhang, Xu, Shen, Li, Zeng, Wei, Cheng, Liu, and Zhou]{skywork-or1-2025}
Jujie He, Jiacai Liu, Chris~Yuhao Liu, Rui Yan, Chaojie Wang, Peng Cheng, Xiaoyu Zhang, Fuxiang Zhang, Jiacheng Xu, Wei Shen, Siyuan Li, Liang Zeng, Tianwen Wei, Cheng Cheng, Yang Liu, and Yahui Zhou.
\newblock Skywork open reasoner series.
\newblock \url{https://capricious-hydrogen-41c.notion.site/Skywork-Open-Reaonser-Series-1d0bc9ae823a80459b46c149e4f51680}, 2025.
\newblock Notion Blog.

\bibitem[Zhang et~al.(2025)Zhang, Lin, Hou, Feng, and Li]{zhang2025adaptthinkreasoningmodelslearn}
Jiajie Zhang, Nianyi Lin, Lei Hou, Ling Feng, and Juanzi Li.
\newblock Adaptthink: Reasoning models can learn when to think, 2025.
\newblock URL \url{https://arxiv.org/abs/2505.13417}.

\bibitem[Tu et~al.(2025)Tu, Lin, Zhang, Tian, Li, Lan, and Zhao]{tu2025learningthinkshapingadaptive}
Songjun Tu, Jiahao Lin, Qichao Zhang, Xiangyu Tian, Linjing Li, Xiangyuan Lan, and Dongbin Zhao.
\newblock Learning when to think: Shaping adaptive reasoning in r1-style models via multi-stage rl, 2025.
\newblock URL \url{https://arxiv.org/abs/2505.10832}.

\bibitem[Chen et~al.(2021)Chen, Tworek, Jun, Yuan, de~Oliveira~Pinto, Kaplan, Edwards, Burda, Joseph, Brockman, Ray, Puri, Krueger, Petrov, Khlaaf, Sastry, Mishkin, Chan, Gray, Ryder, Pavlov, Power, Kaiser, Bavarian, Winter, Tillet, Such, Cummings, Plappert, Chantzis, Barnes, Herbert-Voss, Guss, Nichol, Paino, Tezak, Tang, Babuschkin, Balaji, Jain, Saunders, Hesse, Carr, Leike, Achiam, Misra, Morikawa, Radford, Knight, Brundage, Murati, Mayer, Welinder, McGrew, Amodei, McCandlish, Sutskever, and Zaremba]{chen2021evaluatinglargelanguagemodels}
Mark Chen, Jerry Tworek, Heewoo Jun, Qiming Yuan, Henrique~Ponde de~Oliveira~Pinto, Jared Kaplan, Harri Edwards, Yuri Burda, Nicholas Joseph, Greg Brockman, Alex Ray, Raul Puri, Gretchen Krueger, Michael Petrov, Heidy Khlaaf, Girish Sastry, Pamela Mishkin, Brooke Chan, Scott Gray, Nick Ryder, Mikhail Pavlov, Alethea Power, Lukasz Kaiser, Mohammad Bavarian, Clemens Winter, Philippe Tillet, Felipe~Petroski Such, Dave Cummings, Matthias Plappert, Fotios Chantzis, Elizabeth Barnes, Ariel Herbert-Voss, William~Hebgen Guss, Alex Nichol, Alex Paino, Nikolas Tezak, Jie Tang, Igor Babuschkin, Suchir Balaji, Shantanu Jain, William Saunders, Christopher Hesse, Andrew~N. Carr, Jan Leike, Josh Achiam, Vedant Misra, Evan Morikawa, Alec Radford, Matthew Knight, Miles Brundage, Mira Murati, Katie Mayer, Peter Welinder, Bob McGrew, Dario Amodei, Sam McCandlish, Ilya Sutskever, and Wojciech Zaremba.
\newblock Evaluating large language models trained on code, 2021.
\newblock URL \url{https://arxiv.org/abs/2107.03374}.

\bibitem[Jain et~al.(2024)Jain, Han, Gu, Li, Yan, Zhang, Wang, Solar-Lezama, Sen, and Stoica]{jain2024livecodebenchholisticcontaminationfree}
Naman Jain, King Han, Alex Gu, Wen-Ding Li, Fanjia Yan, Tianjun Zhang, Sida Wang, Armando Solar-Lezama, Koushik Sen, and Ion Stoica.
\newblock Livecodebench: Holistic and contamination free evaluation of large language models for code, 2024.
\newblock URL \url{https://arxiv.org/abs/2403.07974}.

\bibitem[Austin et~al.(2021)Austin, Odena, Nye, Bosma, Michalewski, Dohan, Jiang, Cai, Terry, Le, and Sutton]{austin2021programsynthesislargelanguage}
Jacob Austin, Augustus Odena, Maxwell Nye, Maarten Bosma, Henryk Michalewski, David Dohan, Ellen Jiang, Carrie Cai, Michael Terry, Quoc Le, and Charles Sutton.
\newblock Program synthesis with large language models, 2021.
\newblock URL \url{https://arxiv.org/abs/2108.07732}.

\bibitem[Lightman et~al.(2023)Lightman, Kosaraju, Burda, Edwards, Baker, Lee, Leike, Schulman, Sutskever, and Cobbe]{lightman2023letsverifystepstep}
Hunter Lightman, Vineet Kosaraju, Yura Burda, Harri Edwards, Bowen Baker, Teddy Lee, Jan Leike, John Schulman, Ilya Sutskever, and Karl Cobbe.
\newblock Let's verify step by step, 2023.
\newblock URL \url{https://arxiv.org/abs/2305.20050}.

\bibitem[Rein et~al.(2023)Rein, Hou, Stickland, Petty, Pang, Dirani, Michael, and Bowman]{Rein2023GPQAAG}
David Rein, Betty~Li Hou, Asa~Cooper Stickland, Jackson Petty, Richard~Yuanzhe Pang, Julien Dirani, Julian Michael, and Samuel~R. Bowman.
\newblock Gpqa: A graduate-level google-proof q\&a benchmark.
\newblock \emph{ArXiv}, abs/2311.12022, 2023.
\newblock URL \url{https://api.semanticscholar.org/CorpusID:265295009}.

\end{thebibliography}
